\documentclass[letterpaper]{article} % DO NOT CHANGE THIS
\usepackage{aaai2026}  % DO NOT CHANGE THIS
\usepackage{times}  % DO NOT CHANGE THIS
\usepackage{helvet}  % DO NOT CHANGE THIS
\usepackage{courier}  % DO NOT CHANGE THIS
\usepackage[hyphens]{url}  % DO NOT CHANGE THIS
\usepackage{graphicx} % DO NOT CHANGE THIS
\usepackage{natbib}  % DO NOT CHANGE THIS AND DO NOT ADD ANY OPTIONS TO IT
\usepackage{caption} % DO NOT CHANGE THIS AND DO NOT ADD ANY OPTIONS TO IT
\usepackage{appendix}

\usepackage{appendix}
\usepackage{adjustbox}
\usepackage[ruled,vlined]{algorithm2e}
\usepackage{amsmath}
\usepackage{algpseudocode}
\usepackage{amssymb}
\usepackage{booktabs}
\usepackage{tabularx}
\usepackage{array} % 确保已经导入了 array 包
\usepackage{tabularx} % 如果你正在使用 tabularx 环境，也需要导入它
\usepackage{booktabs}    % 表格横线
\usepackage{multirow}    % \multirow
\usepackage{xcolor}      % \cellcolor
\usepackage{colortbl}    % 表格颜色兼容
\usepackage{subcaption}  % 提供子表格功能
\usepackage{graphicx}
\usepackage{xcolor}
\definecolor{comprehensive}{RGB}{0,128,0}    % 绿色：代表综合
\definecolor{conservative}{RGB}{135,206,235}     % 蓝色：代表保守
\definecolor{aggressive}{RGB}{255,0,0}       % 红色：代表激进
\usepackage{subcaption} % Make sure you include this package in your preamble
\newcolumntype{C}{>{\centering\arraybackslash}X}

\usepackage{newfloat}
\usepackage{listings}
\DeclareCaptionStyle{ruled}{labelfont=normalfont,labelsep=colon,strut=off} % DO NOT CHANGE THIS
\usepackage{floatrow}

\DeclareNewFloatType{listing}{fileext=lol,
   name=Listing,
   placement=tb,
   within=section}

\definecolor{waymogreen}{HTML}{00E89D}
\definecolor{waymolgreen}{HTML}{99F7D7} % alpha=60%
\definecolor{waymollgreen}{HTML}{CCFAEB} % alpha=80%
\definecolor{waymoblue}{HTML}{0077FF}
\definecolor{waymolblue}{HTML}{99B7FF}  % alpha=60%
\definecolor{waymollblue}{HTML}{CCE4FF} % alpha=80%
\definecolor{waymolgray}{HTML}{F0F0F0}  % 浅灰色
\definecolor{mylightgray}{gray}{0.6} % 灰度值
\usepackage{colortbl}       % Cell background colors
\usepackage{pifont}         % Provides \cmark and \xmark
\usepackage{multirow}
\nocopyright
\newcommand{\eqstar}{\textsuperscript{\raisebox{-1.5pt}{\scriptsize *}}}

\title{EvaDrive: Evolutionary Adversarial Policy Optimization for End-to-End Autonomous Driving}
\author{
    Siwen Jiao\textsuperscript{\rm 1,3}\eqstar, Kangan Qian\textsuperscript{\rm 2,3}\eqstar, Hao Ye\textsuperscript{\rm 3}\thanks{Equal contribution.}\thanks{Project leader.}, Yang Zhong\textsuperscript{\rm 3}, Ziang Luo\textsuperscript{\rm 2,3}, Sicong Jiang\textsuperscript{\rm 6}, Zilin Huang\textsuperscript{\rm 7}, Yangyi Fang\textsuperscript{\rm 2}, Jinyu Miao\textsuperscript{\rm 2}, Zheng Fu\textsuperscript{\rm 2}, Yunlong Wang\textsuperscript{\rm 2}, Kun Jiang\textsuperscript{\rm 2}, Diange Yang\textsuperscript{\rm 2}\textsuperscript{‡}, \\Rui Fan\textsuperscript{\rm 4}, Baoyun Peng\textsuperscript{\rm 5}\thanks{Corresponding author.}
}
\affiliations{
    \textsuperscript{\rm 1}National University of Singapore\quad
    \textsuperscript{\rm 2}Tsinghua University\quad
    \textsuperscript{\rm 3}Xiaomi EV             \\ 
    \textsuperscript{\rm 4}Tongji University\quad  
    \textsuperscript{\rm 5}Advanced Institute of Big Data,Beijing   \\
    \textsuperscript{\rm 6}McGill University\quad
    \textsuperscript{\rm 7}University of Wisconsin–Madison \\
    e1349765@u.nus.edu

}

\usepackage{bibentry}
\begin{document}
\maketitle
% Uncomment the following to link to your code, datasets, an extended version or similar.
% You must keep this block between (not within) the abstract and the main body of the paper.
% \begin{links}
%     \link{Code}{https://aaai.org/example/code}
%     \link{Datasets}{https://aaai.org/example/datasets}
%     \link{Extended version}{https://aaai.org/example/extended-version}
% \end{links}

\begin{abstract}
Autonomous driving faces significant challenges in achieving human-like iterative decision-making, which continuously generates, evaluates, and refines trajectory proposals. 
% Existing approaches fall short: modular pipelines suffer from cascading errors; end-to-end models lack multi-modal generalization; and generation-evaluation frameworks isolate trajectory generation from quality assessment, preventing iterative refinement essential for robust planning. Furthermore, prevailing reinforcement learning (RL) approaches collapse multi-dimensional driving preferences (safety, comfort, progress) into scalar rewards, obscuring critical trade-offs and yielding suboptimal policies.
Current generation-evaluation frameworks isolate trajectory generation from quality assessment, preventing iterative refinement essential for planning, while reinforcement learning methods collapse multi-dimensional preferences into scalar rewards, obscuring critical trade-offs and yielding scalarization bias.
% suboptimal policies.
% To overcome these issues, we introduce EvaDrive, an innovative framework achieving a tightly coupled, closed-loop co-evolution between trajectory generation and evaluation through multi-round adversarial optimization. 
To overcome these issues, we present EvaDrive, a novel multi-objective reinforcement learning framework that establishes genuine closed-loop co-evolution between trajectory generation and evaluation via adversarial optimization. 
% EvaDrive frames autonomous driving as a multi-objective reinforcement learning problem, featuring a hierarchical context-aware planner that combines autoregressive intent modeling for temporal causality with diffusion-based refinement for spatial flexibility, enabling precise trajectory generation. It further establishes a closed-loop co-evolution framework linking generation and evaluation, facilitating continuous adaptation via dynamic feedback and cleverly leveraging a Pareto front selection mechanism to avoid local optima. 
EvaDrive frames trajectory planning as a multi-round adversarial game. In this game, a hierarchical generator continuously proposes candidate paths by combining autoregressive intent modeling for temporal causality with diffusion-based refinement for spatial flexibility. These proposals are then rigorously assessed by a trainable multi-objective critic that explicitly preserves diverse preference structures without collapsing them into a single scalarization bias.
% Finally, we propose a multi-round adversarial optimization mechanism where a dynamic interplay between the generator and evaluator continuously improves trajectory quality, effectively escapes local optima, and allows for diverse trajectory styles by dynamically weighting multiple objectives. 
This adversarial interplay, guided by a Pareto frontier selection mechanism, enables iterative multi-round refinement, effectively escaping local optima while preserving trajectory diversity.
% By mimicking human-like iterative decision-making-generating, evaluating, and refining trajectories with adversarial feedback-EvaDrive significantly enhances trajectory quality, robustness, and adaptability in complex dynamic scenarios.
Extensive experiments on NAVSIM and Bench2Drive benchmarks demonstrate SOTA performance, achieving 94.9 PDMS on NAVSIM v1 (surpassing DiffusionDrive by 6.8, DriveSuprim by 5.0, and TrajHF by 0.9) 
and 64.96 Driving Score on Bench2Drive. EvaDrive generates diverse driving styles via dynamic weighting without external preference data, introducing a closed-loop adversarial framework for human-like iterative decision-making, offering a novel scalarization-free trajectory optimization approach.
\end{abstract}

\section{Introduction}

Autonomous driving has made significant advances in recent years~\cite{qian2025agentthink}. These advances enable cars to navigate increasingly complex environments. Despite substantial progress in components such as perception~\cite{qian2024priormotion, qian2025lego}, prediction~\cite{wang2022sti, shi2024streamingflow}, and planning~\cite{fu2025enhancing, qian2024spider}, integrating these capabilities into holistic driving behavior remains challenge. Central to this integration challenge is vehicle trajectory planning, which is the process of generating feasible, safe, and efficient motion trajectories.
% that bridge the gap between high-level navigation objectives and low-level vehicle control.

\begin{figure}[htbp]
  \centering
  \includegraphics[width=0.95\linewidth]{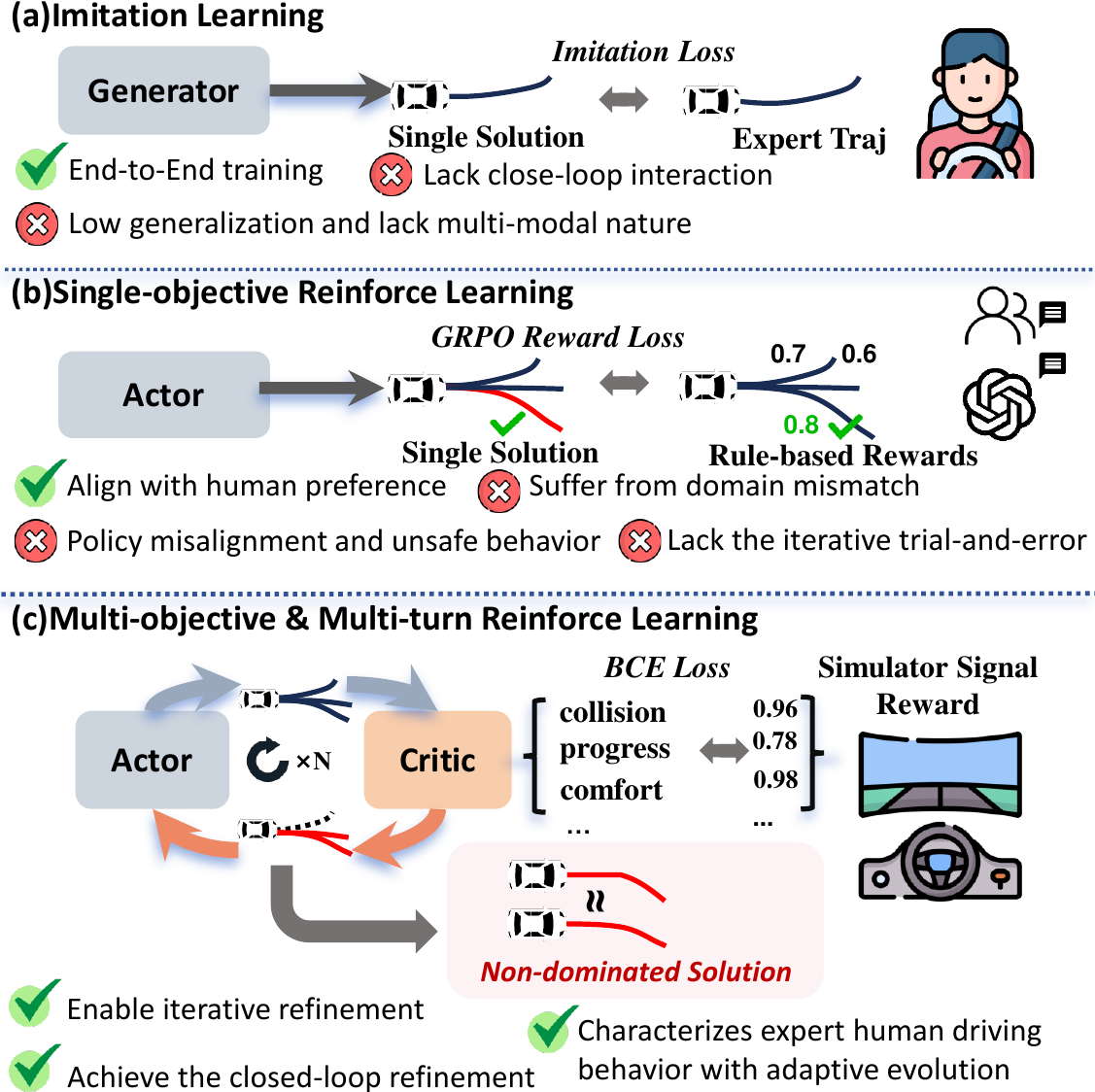}
  % \vspace{-8pt}
  \caption{Comparison of end-to-end paradigms.
(a) Imitation Learning: Mimics expert trajectories via imitation loss.
(b) Single-objective Reinforcement Learning: Uses GRPO reward loss for preference alignment.
(c) Proposed Multi-objective \& Multi-turn Reinforcement Learning: Employs iterative Actor-Critic interaction for closed-loop refinement.}
  \label{fig:motivation}
% \vspace{-10pt}
\end{figure}
Traditional modular pipelines offer interpretability and decomposability, yet suffer from cumulative errors across stages. In contrast, end-to-end approaches~\cite{hu2022st, hu2023planning} enable holistic optimization but often struggle with generalization and fail to model the inherent multimodality in complex driving decisions~\cite{jiang2025survey}.Recently, generation-evaluation frameworks have gained attention for their ability to reconcile diversity with controllability.Trajectory generators capture uncertainty by employing probabilistic models, such as diffusion-based approaches~\cite{yu2024ldp, xing2025goalflow, song2025breaking} to synthesize diverse trajectory candidates. In parallel, evaluators learn differentiable scoring functions to rank these trajectories with respect to predefined multi-objective metrics~\cite{chen2024vadv2, li2024hydra, li2025generalized}.

Despite recent advances, current generation-evaluation frameworks treat planning and assessment as independent, sequential processes, lacking the closed-loop interaction essential to human-like decision-making. In contrast, expert drivers continuously generate, evaluate, and refine actions based on dynamic feedback. To bridge this gap, reinforcement learning (RL) methods have emerged as promising alternatives~\cite{tang2025plan, li2025drive, jiang2025alphadrive}. Approaches like TrajHF~\cite{li2025finetuning} use GRPO to fine-tune trajectory models with human feedback, while DPO-style methods~\cite{rafailov2023direct} optimize behavior via pairwise preference alignment.
However, these methods face key limitations in autonomous driving. First, their reliance on human-annotated pairwise rankings introduces subjectivity and inconsistency, risking policy misalignment~\cite{kiran2021deep}. Second, their origins in language tasks lead to domain mismatch in continuous, high-dimensional driving settings~\cite{zhou2025safemvdrive}. Most critically, one-step optimization lacks the iterative refinement necessary for safe, adaptive planning—limiting robustness in safety-critical environments~\cite{shani2024multi}.

In fact, human drivers continuously evaluate and refine trajectory decisions through internal imagination and interaction with the environment, reasoning over multiple alternatives before committing to action. This observation motivates a key question:
\textit{Can autonomous driving cars leverage multi-modal trajectory proposals and multi-round refinement to enhance planning robustness and align with nuanced human preferences?} We answer affirmatively with \textbf{EvaDrive}, a novel framework that reformulates trajectory planning as a multi-objective reinforcement learning problem with genuine closed-loop interaction.
% between generation and evaluation. 

Unlike prior RL methods that collapse multi-dimensional preferences into scalar rewards, EvaDrive employs a trainable multi-objective reward model that preserves the structure of diverse preferences across safety, comfort, and efficiency without scalarization bias. Our approach is motivated by two key insights: autonomous driving provides naturally quantifiable metrics, avoiding noisy pairwise rankings required by GRPO; and feasible trajectories form a Pareto non-dominated set, where trade-offs across objectives exist. Traditional scalarization compresses these objectives into a single metric, obscuring optimal solutions and reducing policy diversity. EvaDrive uses adversarial co-evolution between a hierarchical generator and multi-objective critic to enable iterative multi-round refinement, escaping local optima while retaining RL’s trial-and-error nature. The main contributions are:

\begin{itemize}
\item \textbf{Hierarchical Planner}: Combines autoregressive intent modeling (to capture temporal causality) and diffusion-based refinement (to provide spatial flexibility), enabling accurate trajectory modeling.

\item \textbf{Multi-turn Optimization  Mechanism}: Connects the trajectory generation and evaluation processes, achieving continuous adaptation through dynamic feedback, and skillfully avoids local optima by leveraging the Pareto frontier selection mechanism.

\item \textbf{Adversarial Policy Optimization}: Through the dynamic game between the generator and evaluator, it not only drives continuous improvement of trajectory quality and effectively escapes local optima, but also enables the generation of trajectories with different styles by adjusting the dynamic weighting ratios of multiple optimization objectives.
% \item \textbf{Adversarial Policy Optimization}: The generator-evaluator dynamic game continuously improves trajectory quality while escaping local optima, with adjustable objective weightings enabling diverse trajectory generation.
\end{itemize}
% As shown in Fig.~\ref{fig:motivation}, this paradigm shift enables the system to achieve the closed-loop refinement and adaptive evolution that characterizes expert human driving behavior. The main contributions of this paper are as follows:
% \begin{itemize}
%     \item Propose a \textbf{closed-loop co-evolution framework} that establishes genuine closed-loop interaction between generation and evaluation components through adversarial training, enabling continuous co-adaptation to overcome the generation-evaluation separation challenges.
    
%     \item Introduce a \textbf{multi-round Pareto optimization} mechanism with Pareto-based selection that systematically escapes local optima while ensuring continuous trajectory quality improvement through iterative refinement.
    
%     \item Present a \textbf{hierarchical context-aware planner} combining autoregressive intent modeling with diffusion-based refinement, enhanced by adversarial preference optimization that maintains multi-objective reward structures for real-time adaptation to dynamic scenarios.
% \end{itemize}

\section{Related Works}
\subsection{E2E AD}
Imitation learning (IL) has been widely adopted for E2E trajectory planning in autonomous driving, where raw sensor inputs are mapped to expert demonstrations~\cite{shao2023safety, qian2024fasionad, qian2025fasionad++}. UniAD~\cite{hu2023planning}, Transfuser~\cite{chitta2022transfuser} leverage Bird-Eye-View (BEV) representations and introduce unified multi-task and safety-aware architectures to build planning-centric frameworks. VAD~\cite{jiang2023vad} further improve this design by employing query-based representations and constructing a trajectory vocabulary, effectively transforming the problem from regression into a discrete classification task.

% To model the interactions between the ego vehicle and other traffic participants, GenAD~\cite{} and PPAD~\cite{} utilize instance-level visual features, while GraphAD~\cite{} represents these features as nodes in a graph structure. SparseDrive~\cite{} propose a fully sparse architecture, achieving greater efficiency and superior planning performance. However, most of these methods typically rely on constructing heavy BEV features, which prevents downstream tasks from learning directly from raw sensor data.

% To address this issue, PARA-Drive~\cite{} and DriveTransformer~\cite{} introduce a parallel pipeline, enabling a more unified and scalable framework for end-to-end driving systems. Additionally, methods such as AD-MLP~\cite{} adopt a simple network architecture and are tested on nuScenes~\cite{}, relying solely on the ego vehicle's status information. These approaches highlight the potential over-reliance on ego status in simpler driving scenarios while also revealing limitations in the effective use of perception data.
Recent benchmarks like NavSim~\cite{dauner2024navsim} advance E2E planning with multi-agent scenarios, while frameworks including Hydra-MDP~\cite{li2024hydra}, GTRS~\cite{li2025generalized} and iPad~\cite{guo2025ipad} employ scorer-based trajectory synthesis integrating environmental interactions. However, these universally lack generator-scorer feedback loops. Contrastingly, our human decision-inspired solution integrates trajectory generation and evaluation within an iterative optimization cycle.

\subsection{RL in generative task(multi-object and multi-turn)}
%主要讲多轮和多目标这两个, AI找的文献，幻觉感觉很大的
% Chen et al. (2023). Memory-Enhanced Dialogue Policy for Multi-Turn Interaction. ACL.
% Feng et al. (2020). Curriculum Learning for Dialogue Policy Optimization. NeurIPS.
% Guo et al. (2021). Actor-Critic with Memory Networks for Long-Horizon Dialogues. EMNLP.
% He et al. (2022). Inverse RL for Preference-Aware Dialogue Systems. ICML.
% Kostrikov et al. (2021). Offline Reinforcement Learning for Dialogue. ICLR.
% Li et al. (2020). Hierarchical RL for Multi-Objective Dialogues. AAAI.
% Ouyang et al. (2022). Training Language Models to Follow Instructions with RL. NeurIPS.
% Wang et al. (2019). Weighted Sum Methods in Dialogue Policy Optimization. ACL.
% Zhang et al. (2021). Pareto-Based Multi-Objective Dialogue Generation. EACL.
% Zhao et al. (2022). Hierarchical Dialogue Policy Optimization. COLING.
RL has emerged as a critical paradigm for optimizing generative models in complex, goal-driven scenarios, particularly in multi-objective and multi-turn dialogue systems\cite{ouyang2022training, zhou2025safemvdrive, shani2024multi}. These tasks require balancing competing objectives (e.g., fluency, coherence, task success) and maintaining long-term context consistency across multiple interaction turns.

% Key challenges include reward engineering for non-differentiable objectives, exploration-exploitation trade-offs in sparse reward settings, and generalization across diverse user behaviors. Emerging directions focus on integrating RL with large language models (LLMs) for reward modeling\cite{ouyang2022training} and leveraging offline RL to reduce real-time interaction costs\cite{gupta2023offline}. Hybrid approaches combining RL with pre-trained dialogue systems (e.g., DialoGPT\cite{yang2021ubar}) show promise in achieving both task efficiency and natural language fluency. In autonomous driving, reinforcement learning (RL) methods like GRPO~\cite{shao2024deepseekmath} and DPO~\cite{rafailov2023direct} are typically deployed during post-training to align trajectories with human preferences. By contrast, our framework employs \textit{multi-round multi-objective RL} with APO method~\cite{cheng2023adversarial} for trajectory optimization.
Key challenges include non-differentiable reward engineering, sparse-reward exploration-exploitation tradeoffs, and cross-user generalization. Emerging solutions integrate LLMs for reward modeling~\cite{ouyang2022training} and offline RL to reduce interaction costs~\cite{gupta2023offline}, while hybrid methods with pretrained dialogue systems~\cite{yang2021ubar} balance task efficiency with natural fluency. In autonomous driving, post-training alignment methods like GRPO~\cite{shao2024deepseekmath} and DPO~\cite{rafailov2023direct} optimize human preference matching. Contrastingly, our framework pioneers multi-round multi-objective RL with Adversarial Policy Optimization (APO)~\cite{cheng2023adversarial} for trajectory optimization.

\begin{figure*}[htbp]
  \centering
  \includegraphics[width=0.98\linewidth]{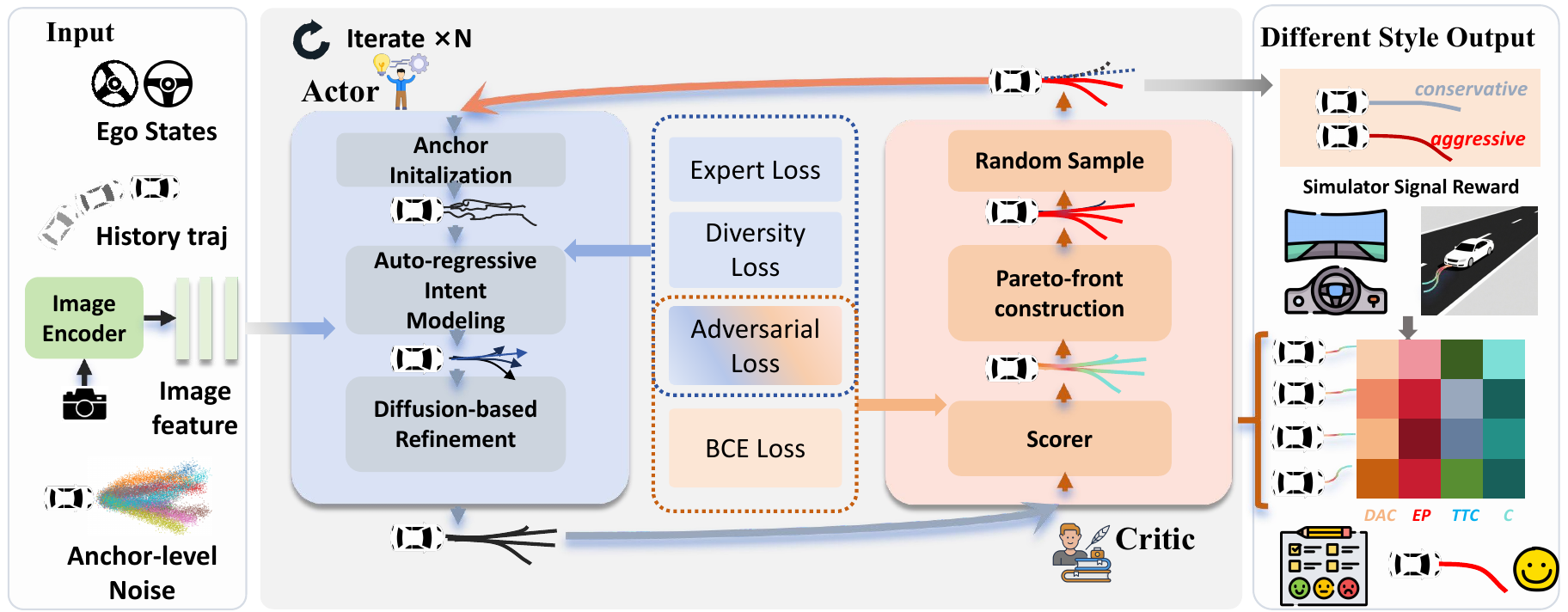}  
  % \vspace{-8pt}
\caption{Overall framework of \textit{EvaDrive}. Our method generates and evaluates trajectories in an end-to-end, multi-objective manner. (a) The Actor performs anchor initialization, intent modeling, and diffusion-based refinement to propose diverse trajectory candidates. (b) The Critic evaluates them using multi-objective simulator feedback and constructs a Pareto front. The best candidates are iteratively reused to guide future sampling, enabling style-adaptive and preference-aware planning.}

  \label{fig:framework}
  % \vspace{-15pt}
\end{figure*}

\section{Problem Formulation}

% \subsection{Reinforcement Learning for Autonomous Driving}
% Reinforcement Learning (RL) provides a framework for sequential decision-making, wherein an agent learns a policy \(\pi_\theta(a \mid s)\) that maximizes the expected cumulative reward:
% \begin{equation}
% J(\theta) = \mathbb{E}_{s \sim \mathcal{D},\, a \sim \pi_\theta} \left[ R(s, a) \right],
% \end{equation}
% where \(s\) denotes the state, \(a\) the action, and \(R(s,a)\) a scalar reward indicating action quality. In actor-critic paradigms, the policy \(\pi_\theta\) serves as the "actor" that proposes actions, while the reward function acts as a "critic" to evaluate them. In autonomous driving settings, \(\pi_\theta\) functions as a trajectory planner: \(s\) encodes the driving context (including perception, traffic information, and prior decisions), \(a\) represents the ego vehicle's future trajectory, and the scalar reward \(R(s,a)\) is replaced by a domain-specific driving reward \(R_{\text{drive}}(s,a)\), reflecting safety, comfort, and rule compliance.
\subsection{Reinforcement Learning for Autonomous Driving}
Reinforcement Learning (RL) optimizes a policy \(\pi_\theta(a \mid s)\) to maximize expected cumulative reward:
\begin{equation}
J(\theta) = \mathbb{E}_{s \sim \mathcal{D},\, a \sim \pi_\theta} \left[ R(s, a) \right],
\end{equation}
where \(s\) and \(a\) denote state and action, respectively. In actor-critic frameworks, \(\pi_\theta\) proposes actions (actor), and a reward function evaluates them (critic). For autonomous driving, \(\pi_\theta\) plans future trajectories, with \(s\) encoding scene context and \(a\) representing the ego trajectory. The reward \(\mathbf{R_{\text{drive}}}(s,a)\) reflects driving-specific objectives like safety, comfort, and traffic rule compliance.

\subsection{Multi-Objective Optimization}
Conventional RL approaches (e.g., DPO, GRPO) simplify complex decision-making to scalar reward maximization, insufficient for autonomous driving—which requires joint optimization of conflicting objectives: safety, comfort, efficiency, and rule adherence.

To capture this multi-faceted nature, we represent the driving reward as a \textbf{multi-objective vector}:
\[
\mathbf{R_{\text{drive}}}(s,a) = \big[ r_1(s,a),\; r_2(s,a),\; \dots,\; r_K(s,a) \big]^\top \in \mathbb{R}^K,
\]
where each \(r_i(s,a)\) corresponds to a distinct objective. The learning problem becomes maximizing the expected vector-valued return:
\begin{equation}
J(\theta) = \mathbb{E}_{(s,a) \sim \pi_\theta} \left[ \mathbf{R_{\text{drive}}}(s,a) \right] \in \mathbb{R}^K.
\end{equation}

For conflicting objectives (e.g., safety vs. efficiency), scalarization masks key trade-offs. Instead, we seek \textbf{Pareto-optimal solutions}—no other solution improves any objective without degrading at least one other—forming the target \textbf{Pareto frontier} for multi-objective optimization.

\subsection{Reintroducing the Reward Model}
To avoid limitations of implicit or oversimplified rewards, we adopt a reward learning approach akin to RLHF, introducing a learnable reward model \(r_\phi\) that predicts \(\mathbf{R_{\text{drive}}}(s,a)\), trained as follows:

\begin{itemize}
    \item \textbf{Preference Data:} Ground-truth \(\mathbf{R_{\text{drive}}}(s,a)\) derive from simulation-based trajectory metrics (e.g., obstacle proximity for safety, jerk for comfort).
    
    \item \textbf{Reward Model Training:} \(r_\phi\) is supervised to match ground truth via MSE:
    \begin{equation}
    \phi^* = \arg\min_{\phi} \mathbb{E}_{(s,a) \sim \mathcal{D}} \left[ \left\| r_\phi(s,a) - \mathbf{R_{\text{drive}}}(s,a) \right\|^2 \right],
    \end{equation}
    where \(r_\phi(s,a) \in \mathbb{R}^K\) is the predicted reward vector.
    
    \item \textbf{Policy Optimization:} Trained \(r_\phi\) serves as a differentiable proxy for environment feedback, enabling policy updates via multi-objective signals.
\end{itemize}

%TODO:unify the notion system
%统一符号系统！尤其是张量形状和角标

\section{Method}
Within the framework of multi-objective reinforcement learning, we now detail the core components of our approach, including the trajectory generation and evaluation modules (Actor and Critic), the multi-turn optimization mechanism, and the Adversarial Preference Optimization paradigm.

\subsection{Trajectory Planner as Actor}

\begin{figure}[h!]
  \centering
  \includegraphics[width=0.98\linewidth]{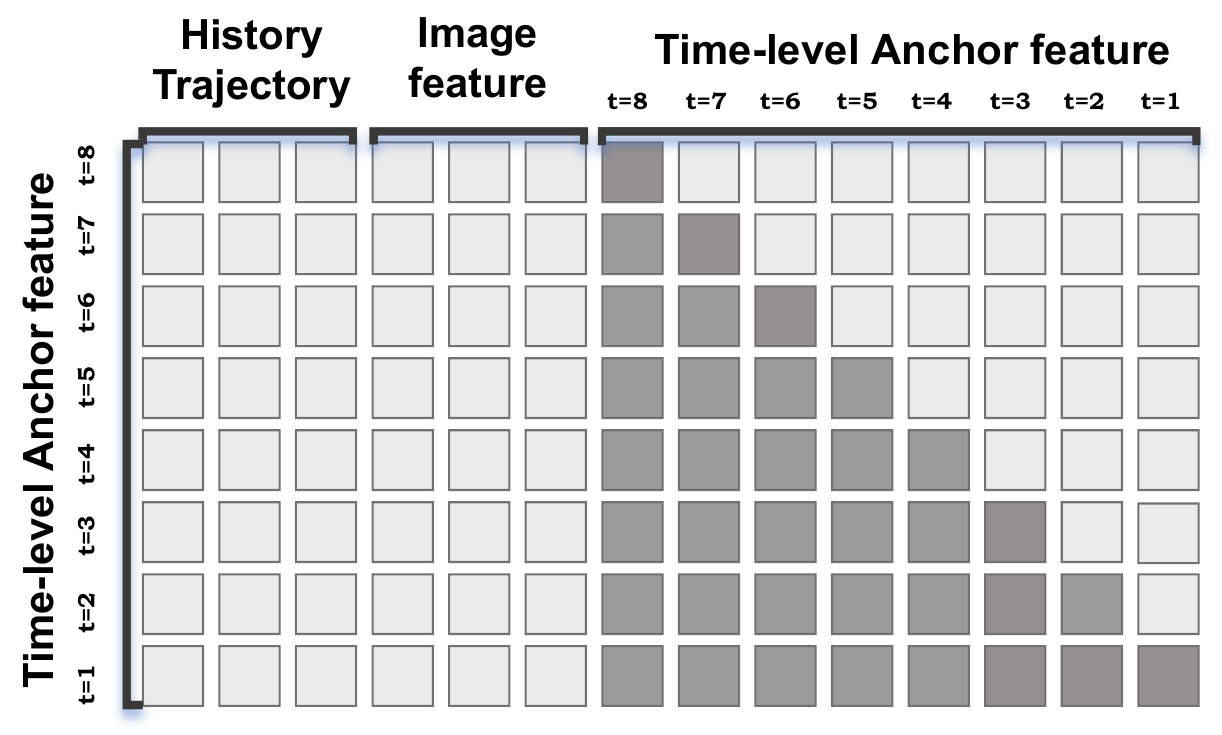}
  % \vspace{-8pt}
    \caption{Attention Mask of Autoregressive Intent Generator, grey: visible elements, black: invisible elements}
  \label{fig:actor_architecture}
  % \vspace{-15pt}
\end{figure}
We propose a lightweight structured trajectory planner, termed hierarchical modeling planner, serving as the \textbf{Actor} in our reinforcement learning framework. This Actor integrates two core components: an \textit{autoregressive intent generator} and a \textit{diffusion-based refiner}, aiming to address two critical challenges in planning: temporal causality modeling and refined optimization of global spatial trajectories under uncertainty. First, the current ego state \(\mathbf{s}_{\text{ego}}\) is extracted, based on which a Multi-Layer Perceptron (MLP) generates an initial set of candidate sequences \(\mathbf{A}_0 \in \mathbb{R}^{N \times P \times 3}\) (where N denotes the number of sequences, P the number of poses per sequence, and 3 corresponds to xy-coordinates and yaw angle \(\theta\) of each pose). Image features \(\mathbf{F}_{\text{img}} \in \mathbb{R}^{H \times W \times D_f}\) (with \(H \times W\) as spatial resolution and \(D_f\) as feature dimension) are extracted via a frozen visual backbone, serving as global environmental context.
\subsubsection{Stage I: Autoregressive Intent Modeling}
The first stage models temporally coherent motion intent using a \textbf{Multi-Head Cross-Attention (MHCA)} mechanism, denoted as \(\text{MHCA}_T\) to emphasize its temporal specificity.Formally, the query vector \(\mathbf{Q}\) directly adopts the initial candidate sequences \(\mathbf{A}_0\); the key vector \(\mathbf{K}\) and value vector \(\mathbf{V}\) are constructed by concatenating contextual features, all projected to a unified dimension via learnable linear transformations:
\begin{equation}
\mathbf{K} = \mathbf{V} = \text{concat}(\mathbf{A}_0, \mathbf{H}_{hist}, \mathbf{F}_{img})
\end{equation}
where \(\mathbf{H}_{\text{hist}} \in \mathbb{R}^{M \times D_h}\) represents historical trajectory features (M as the number of historical timesteps, \(D_h\) as feature dimension). The temporal cross-attention operation is defined as:
\begin{equation}
\mathbf{A}_{\text{AR}} = \text{MHCA}_T\left(\mathbf{A}_0, \mathbf{K}, \mathbf{V}\right)
\end{equation}
% where \(\mathbf{A}_{\text{AR}} \in \mathbb{R}^{T \times D_p}\) denotes intent features optimized in the temporal dimension, serving as autoregressive candidates for subsequent refinement.A critical design is the rectangular attention mask matching the first dimension of \(\mathbf{Q}\) (i.e., \(\mathbf{A}_0\)) with size T and that of \(\mathbf{K}\) with size \(T + M + H \times W\). This mask enforces temporal causality by restricting each candidate timestep in \(\mathbf{Q}\) to attend only to historical and current elements in \(\mathbf{K}\), which is crucial for learning time-dependent motion patterns (e.g., deceleration before turning) and maintaining the causal order of sequential actions.
where \(\mathbf{A}_{\text{AR}} \in \mathbb{R}^{T \times D_p}\) represents temporally optimized intent features as autoregressive candidates for refinement. A key design is the rectangular attention mask aligning \(\mathbf{Q}\) (of size \(T\)) with \(\mathbf{K}\) (of size \(T + M + H \times W\)), enforcing temporal causality by allowing each timestep in \(\mathbf{Q}\) to attend only to current and past elements in \(\mathbf{K}\). This constraint is essential for learning time-dependent motion patterns (e.g., deceleration before turning) and preserving causal action order.

\subsubsection{Stage II: Diffusion-Based Refinement}
Although the autoregressive module captures temporal coherence, it lacks capabilities in uncertainty modeling and spatial flexibility. To address this, forward noise is injected into \(\mathbf{A}_{\text{AR}}\) using the Denoising Diffusion Implicit Model (DDIM), generating stochastic inputs \(\tilde{\mathbf{A}}^{(t)}\) at diffusion step t:
\begin{equation}
\tilde{\mathbf{A}}^{(t)} = \alpha_t \mathbf{A}_{\text{AR}} + \sigma_t \boldsymbol{\epsilon}, \quad \boldsymbol{\epsilon} \sim \mathcal{N}(0, \mathbf{I}), \quad \tilde{\mathbf{A}}^{(t)} \in \mathbb{R}^{N \times P \times 3}
\end{equation}
where \(\alpha_t\) and \(\sigma_t\) are precomputed DDIM coefficients, and \(\boldsymbol{\epsilon}\) is a standard normal noise vector. These noisy candidates are refined via a second-stage module comprising spatial cross-attention \(\text{MHCA}_{\text{S}}\) followed by a lightweight Transformer decoder \(\mathcal{D}_\theta\). The spatial cross-attention operation is defined as:
\begin{equation}
\mathbf{A}_S = \text{MHCA}_{\text{S}}(\tilde{\mathbf{A}}^{(t)}, \mathbf{F}_{img})
\end{equation}
where \(\mathbf{A}_S \in \mathbb{R}^{N \times P \times 3}\) denotes spatially augmented features. The decoder outputs denoised candidate features:
\begin{equation}
\hat{\mathbf{A}} = \mathcal{D}_\theta(\mathbf{A}_S), \quad \hat{\mathbf{A}} \in \mathbb{R}^{N \times P \times 3}
\end{equation}The final trajectory is decoded via an MLP:
\begin{equation}
\mathbf{T}_{\text{pred}} = \text{MLP}(\hat{\mathbf{A}}), \quad \mathbf{T}_{\text{pred}} \in \mathbb{R}^{N \times P \times 3}
\end{equation}Benefiting from the guidance of autoregressive features, our denoiser enables efficient trajectory generation through single-step denoising, eliminating the need for multi-round inference in traditional diffusion models. This significantly improves computational efficiency, meeting the latency constraints of real-time deployment.
\subsection{Reward Model as Critic}

For reward-guided learning, we introduce a \textbf{reward model as critic} to evaluate predicted trajectories and provide multi-objective feedback for the planner. Unlike traditional scalar reward functions, it offers structured high-dimensional supervision, matching driving's multi-objective nature.

Given the $i$-th predicted trajectory $\mathbf{T}_{\text{pred}}^{(i)}$, we first aggregate its temporal planning features (such as the denoised anchor features $\hat{\mathbf{A}}^{(i)}$ output by the second stage) via temporal max pooling:
\begin{equation}
\mathbf{v}^{(i)} = \text{MaxPool}_t(\hat{\mathbf{A}}^{(i)}), \quad \mathbf{v}^{(i)} \in \mathbb{R}^{D_p}.
\end{equation}
The pooled $\mathbf{v}^{(i)}$ captures the global semantic intent of the trajectory, feeding a MLP that outputs $N$ scalar reward components. The vector composed of these components from all trajectories collectively forms the driving reward vector $\mathbf{R}_{\text{drive}}^{(i)}$ defined earlier:
\begin{equation}
\mathbf{R}_{\text{drive}}^{(i)} = \text{MLP}(\mathbf{v}^{(i)}), \quad \mathbf{R}_{\text{drive}}^{(i)} \in \mathbb{R}^{N}.
\end{equation}

Notably, trajectory evaluation is multi-objective: the critic estimates each objective independently (not as a single scalar), preserving reward diversity, avoiding premature aggregation, and enabling effective Pareto front exploration in subsequent optimization.

\subsection{Multi-Turn Optimization Mechanism}
\label{sec:MTO}
% While we have already defined the multi-objective reward function $\mathbf{R}_{\text{drive}}$, traditional one-shot optimization employed by planning methods in autonomous driving—where policies maximize immediate rewards through single-step decision-making—despite its efficiency, suffers from unidirectionality and the lack of feedback mechanisms. For safety-critical autonomous driving tasks, such a paradigm fails to support sufficient trial-and-error and self-correction, thereby posing challenges to task reliability and safety.
Traditional single-step optimization in autonomous driving planning maximizes immediate rewards but lacks feedback loops. While efficient, this unidirectional approach fails to support iterative refinement through trial-and-error, compromising reliability in safety-critical tasks despite a well-defined reward function $\mathbf{R}_{\text{drive}}$.

To address this, we propose a \emph{multi-turn trajectory optimization mechanism} (Algorithm~\ref{alg:multi-turn-optimization}), which extends the existing multi-objective optimization framework into an iterative reinforcement-style process. Specifically, we define a total of $K$ optimization rounds, indexed by $t = 0, 1, \dots, K-1$. In each round, the actor policy $\pi_\theta$ generates a set of candidate trajectories conditioned on the current planning state $s_t$ and, if applicable, guidance trajectories from the previous round. After completing all $K$ rounds, we select the final output trajectory $a_{\text{final}} \in \mathcal{A}_{K-1}$ from the last candidate set.

We define the overall training objective by evaluating the final output trajectory using the multi-objective reward:
\begin{equation}
J_{\text{multi-turn}}(\theta) = \mathbb{E}_{\mathcal{M}, \, a_{\text{final}} \sim \pi_\theta} \left[ \mathbf{R}_{\text{drive}}(s_{\text{final}}, a_{\text{final}}) \right],
\end{equation}
where $s_{\text{final}}$ denotes the final planning state and $a_{\text{final}}$ is selected from the candidate set $\mathcal{A}_{K-1}$. Here, $\mathbf{R}_{\text{drive}}$ denotes a task-level evaluation function that measures the overall quality of a completed trajectory in the final context.

During training, however, we require a more fine-grained and intermediate assessment of candidate trajectories at each round. To this end, we define a round-level multi-objective reward function $\mathbf{r}(\cdot)$, which shares the same dimensionality as $\mathbf{R}_{\text{drive}}$ but reflects local, per-round trajectory quality. This function acts as a vector-valued critic that outputs $N$ scalar reward components per trajectory.

At each round $t$, the optimization process proceeds as follows:

First, the actor generates a set of candidate trajectories $\mathcal{A}_t = \{a_i\}$ based on the current planning context. For each trajectory $a \in \mathcal{A}_t$, the critic model evaluates its multi-dimensional reward vector:
\begin{equation}
\mathbf{r}(a) = \left[r^{(1)}(a), \dots, r^{(N)}(a)\right],
\end{equation}
where $N$ is the number of reward dimensions (e.g., safety, efficiency, comfort). The scalar functions $r^{(i)}(\cdot)$ represent the $i$-th reward component. Since these objectives often exhibit trade-offs, we avoid reward scalarization to preserve diversity in candidate behavior.

Instead, we extract the Pareto front $\mathcal{P}_t$ from $\mathcal{A}_t$, defined as the set of all non-dominated trajectories:
\begin{equation}
\mathcal{P}_t = \left\{ a_i \in \mathcal{A}_t \mid \nexists a_j \in \mathcal{A}_t,\ \mathbf{r}(a_j) \succ \mathbf{r}(a_i) \right\},
\end{equation}
where $\mathbf{r}(a_j) \succ \mathbf{r}(a_i)$ indicates that trajectory $a_j$ dominates $a_i$, i.e., it performs at least as well in all reward dimensions and strictly better in at least one.

To promote exploration and mitigate overfitting to narrow solutions, we uniformly sample $M$ guidance trajectories from the Pareto front:
\begin{equation}
\left\{ \tilde{a}_t^{(m)} \right\}_{m=1}^{M} \sim \text{UniformSample}(\mathcal{P}_t),
\end{equation}
which are used to condition the actor policy in the next round:
\begin{equation}
\mathcal{A}_{t+1} \leftarrow \pi_\theta(s_{t+1}, \left\{ \tilde{a}_t^{(m)} \right\}_{m=1}^M).
\end{equation}

Optionally, the planning state $s_{t+1}$ can be updated based on the selected trajectory or a state transition function, enabling adaptive planning in dynamic environments. This mechanism enables principled multi-objective planning via Pareto-guided sampling and iterative policy refinement, supporting real-world autonomous driving.

%\vspace{-10pt}

\begin{algorithm}[ht]
\caption{Multi-Turn Trajectory Optimization}
\label{alg:multi-turn-optimization}
\SetAlgoVlined
\SetCommentSty{gray}
\KwIn{Initial planning state $s_0$, actor policy $\pi_\theta$, reward function $\mathbf{r}(\cdot)$, total rounds $K$, sample size $M$}
\KwOut{Final output trajectory $a_{\text{final}} \in \mathcal{A}_{K-1}$}

\For{$t = 0$ \KwTo $K{-}1$}{
    \textbf{1. Generate candidate trajectories} \\
    \eIf{$t = 0$}{
        $\mathcal{A}_0 \leftarrow \pi_\theta(s_0)$\;
    }{
        $\mathcal{A}_t \leftarrow \pi_\theta(s_t, \{\tilde{a}_{t-1}^{(m)}\}_{m=1}^{M})$\;
    }

    \textbf{2. Evaluate reward vectors for each trajectory} \\
    \ForEach{$a \in \mathcal{A}_t$}{
        $\mathbf{r}(a) = [r^{(1)}(a), \dots, r^{(N)}(a)]$\;
    }

    \textbf{3. Extract Pareto front from $\mathcal{A}_t$} \\
    $\mathcal{P}_t \leftarrow \left\{ a_i \in \mathcal{A}_t \,\middle|\, \nexists a_j \in \mathcal{A}_t,\ \mathbf{r}(a_j) \succ \mathbf{r}(a_i) \right\}$\;

    \textbf{4. Sample guidance trajectories from $\mathcal{P}_t$} \\
    $\{\tilde{a}_t^{(m)}\}_{m=1}^{M} \sim \text{UniformSample}(\mathcal{P}_t)$\;

    \textbf{5. Update planning state (if applicable)} \\
    $s_{t+1} \leftarrow \text{UpdateState}(s_t, a_t)$\;
}

\textbf{6. Select final output trajectory} \\
$a_{\text{final}} \in \mathcal{A}_{K-1}$\;  \tcp*[f]{based on $\mathbf{R}_{\text{drive}}$ evaluation}

\Return $a_{\text{final}}$\;
\end{algorithm}

\subsection{Adversarial Policy Optimization (APO)}

To enable preference-aware trajectory optimization, we propose APO, a learning paradigm that frames policy training as a multi-objective optimization problem with adversarial reward learning. This approach draws inspiration from adversarial training principles while specifically addressing the multi-dimensional nature of autonomous driving evaluation tasks. Given a trajectory policy \(\pi_\theta\) and a learnable multi-objective reward model \(r_\phi\) that outputs a reward vector:
\begin{equation}
\mathbf{R}_{\text{drive}}(s,a) = [r_\phi^{(1)}(s,a), \ldots, r_\phi^{(N)}(s,a)]^\top \in \mathbb{R}^N
\end{equation}
where \(N\) denotes the number of distinct driving performance metrics (e.g., safety, comfort, efficiency), we define the core objective of APO. Let \(\mathcal{D}_{\text{expert}} = \{(s_j, a_j)\}_{j=1}^M\) denote the expert demonstration dataset. For each objective dimension \(i \in \{1, \ldots, N\}\), we define:
\begin{equation}
V^{(i)}(\phi, \theta) = \mathbb{E}_{(s,a) \sim \pi_\theta} [r_\phi^{(i)}(s,a)] - \mathbb{E}_{(s,a) \sim \mathcal{D}_{\text{expert}}} [r_\phi^{(i)}(s,a)].
\end{equation}

The overall optimization is a vector-valued objective:
\begin{equation}
\min_{\phi} \max_{\theta} \mathbf{V}(\phi, \theta) = [V^{(1)}(\phi, \theta), \ldots, V^{(N)}(\phi, \theta)]^\top.
\end{equation}

Here, the policy generator (\(\max_\theta\)) aims to produce trajectories that maximize the reward predictions across all dimensions, while the reward model (\(\min_\phi\)) learns to assign higher rewards to expert demonstrations than to generated trajectories, providing gradients that guide the generator toward expert-like behavior.

\subsubsection{Multi-Objective Optimization with Weighted Scalarization}
The optimization addresses a multi-objective problem where each reward component \(r_\phi^{(i)}(s,a)\) corresponds to a distinct driving performance metric. To make the problem computationally tractable, we employ weighted scalarization\cite{zadeh1963optimality} using a preference vector \(\mathbf{w} = \{w_i\}_{i=1}^N\) with \(\sum_{i=1}^N w_i = 1\) and \(w_i \geq 0\). The scalarized objective:
\begin{equation}
\min_{\phi} \max_{\theta} V_{\mathbf{w}}(\phi, \theta) = \sum_{i=1}^N w_i \cdot V^{(i)}(\phi, \theta).
\end{equation}
Different choices of \(\mathbf{w}\) enable the generation of trajectories with distinct behavioral characteristics.
% (e.g., safety-oriented, efficiency-oriented, comfort-oriented).

\subsubsection{Generator Optimization (\(\max_\theta V_{\mathbf{w}}\))}
The generator is updated using the policy gradient:
\begin{equation}
\nabla_\theta V_{\mathbf{w}}(\phi, \theta) = \sum_{i=1}^N w_i \cdot \mathbb{E}_{(s,a) \sim \pi_\theta} \left[ \nabla_\theta \log \pi_\theta(a|s) \cdot r_\phi^{(i)}(s,a) \right],
\end{equation}
with the parameter update:
\begin{equation}
\theta \leftarrow \theta + \eta_\theta \cdot \nabla_\theta V_{\mathbf{w}}(\phi, \theta).
\end{equation}
This weighted approach converts the multi-objective optimization into a single-objective problem while enabling diverse trajectory styles via \(\mathbf{w}\)\cite{marler2004survey}.

\subsubsection{Reward Model Optimization (\(\min_\phi V_{\mathbf{w}}\))}
The reward model is trained to distinguish between generated and expert trajectories by minimizing:
\begin{equation}
\begin{split}
\nabla_\phi V_{\mathbf{w}}(\phi, \theta) &= \sum_{i=1}^N w_i \cdot \bigg[ \mathbb{E}_{(s,a) \sim \pi_\theta} [\nabla_\phi r_\phi^{(i)}(s,a)] \\
&\quad - \mathbb{E}_{(s,a) \sim \mathcal{D}_{\text{expert}}} [\nabla_\phi r_\phi^{(i)}(s,a)] \bigg],
\end{split}
\end{equation}
with the parameter update:
\begin{equation}
\phi \leftarrow \phi - \eta_\phi \cdot \nabla_\phi V_{\mathbf{w}}(\phi, \theta).
\end{equation}
This ensures the reward model maintains balanced discriminative capability across all objectives, preventing the generator from exploiting any single dimension.

\begin{figure*}[!h]
    \centering
    % 左侧子图：添加\phantomcaption激活计数器，确保label有效
    \begin{subfigure}[t]{0.75\linewidth}
        \centering
        \adjustbox{valign=t}{\includegraphics[width=\linewidth]{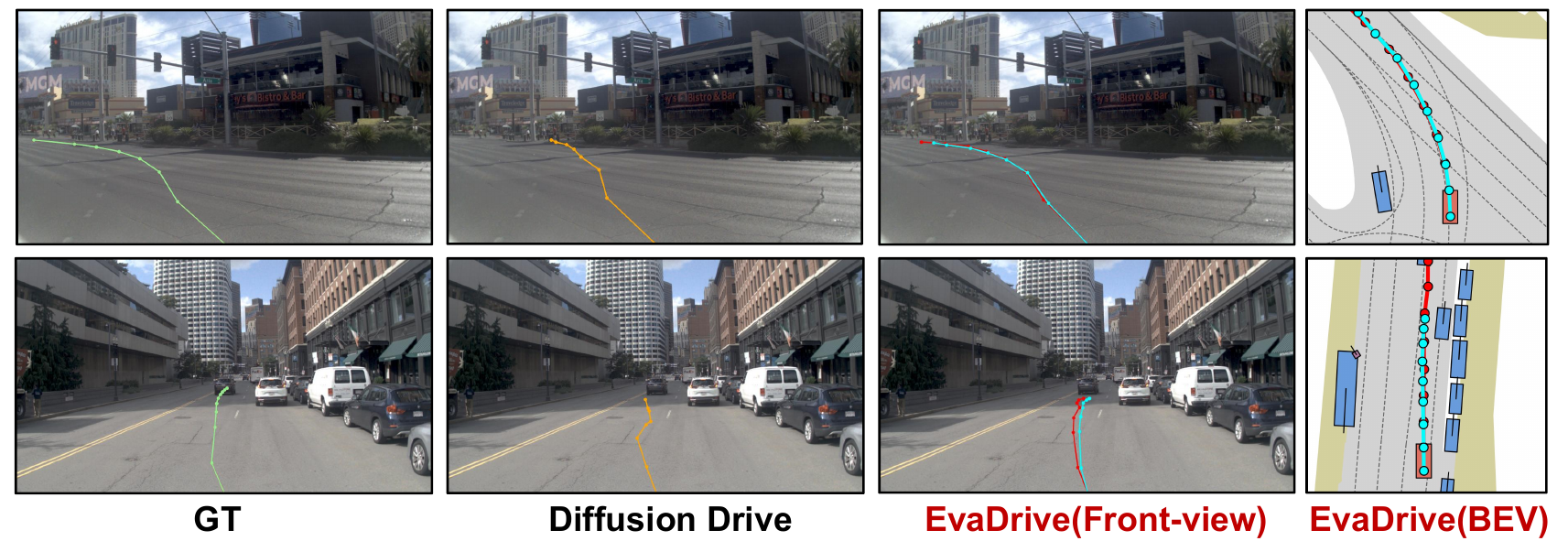}}
        \phantomcaption  % 关键：隐藏子图标题但生成计数器，关联label
        \label{fig:vis1}  % 现在label会正确关联到子图1
    \end{subfigure}%
    \hfill
    % 右侧子图：同样处理
    \begin{subfigure}[t]{0.23\linewidth}
        \centering
        \adjustbox{valign=t}{\includegraphics[width=\linewidth]{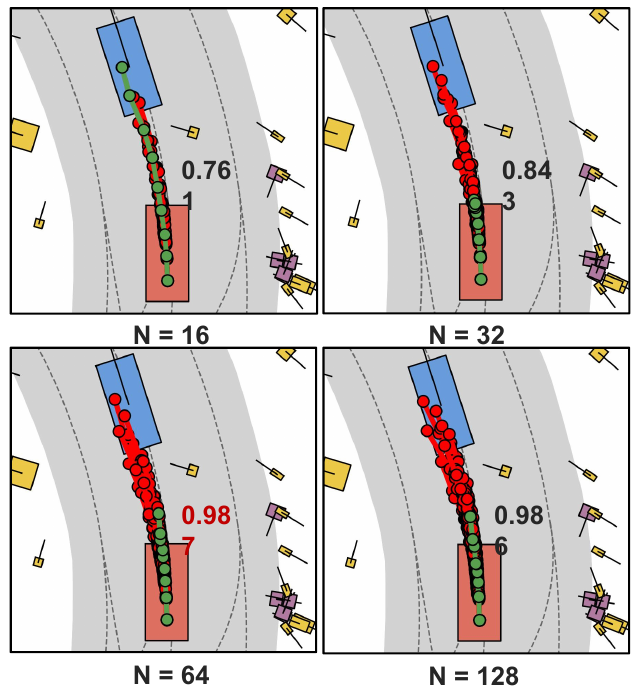}}
        \phantomcaption  % 隐藏标题但激活计数器
        \label{fig:vis2}  % 关联到子图2
    \end{subfigure}
    % 主标题整合子图说明
    \caption{Visualization of end-to-end planning trajectories on Front-view and Bird-eye-view. 
    Left (\ref{fig:vis1}): Qualitative Comparison of Ground Truth, DiffusionDrive, and EvaDrive on NAVSIM NavTest Split (For EvaDrive: \textcolor{aggressive}{red} denotes the aggressive version; \textcolor{conservative}{cyan} denotes our conservative version). 
    Right (\ref{fig:vis2}): Anchor visualization in each iteration (num=64): Red: constructed Pareto solution sets, gray: other eliminated anchors.}
    \label{fig:combined_planning_stages}  % 主图label
\end{figure*}
\subsection{Training Objectives}

The total training loss drives adversarial interaction and joint optimization of the generator (Actor $\pi_\theta$) and multi-objective reward model (Critic $r_\phi$):

\begin{equation}
L = L_{\text{actor}} + L_{\text{critic}}
\end{equation}

This loss formulation enables the Actor to generate high-quality, preference-aligned trajectories while the Critic refines its reward evaluation, fostering their collaborative evolution. Detailed component definitions are in the appendix.

\begin{table}[t!]
\centering
% 设置9号字体（字体大小9pt，行间距10.8pt）
\fontsize{9pt}{10.8pt}\selectfont
\renewcommand{\arraystretch}{1.1} % 保持原有行间距
\setlength{\tabcolsep}{1.5pt} % 进一步压缩列间距以适配单栏（列数较多，适当减小）
% 移除resizebox，避免字体非可控缩放
\begin{tabular}{c *{6}{c}} % 所有列居中对齐
\toprule
\textbf{Method} & \textbf{NC} & \textbf{DAC} & \textbf{EP} & \textbf{TTC} & \textbf{C} & \textbf{PDMS} \\
\midrule
% 第一组方法
Transfuser†\cite{chitta2022transfuser}& 97.7 & 92.8 & 79.2 & 92.8 & 100 & 84.0 \\
UniAD†\cite{hu2023planning} & 97.8 & 91.9 & 78.8 & 92.9 & 100 & 83.4 \\
PARA-Drive†\cite{weng2024drive} & 97.9 & 92.4 & 79.3 & 93.0 & 99.8 & 84.0 \\
VADv2†\cite{jiang2023vad} & 97.9 & 91.7 & 77.6 & 92.9 & 100 & 83.0 \\
LAW†\cite{li2024enhancing} & 96.4 & 95.4 & 81.7 & 88.7 & 99.9 & 84.6 \\
DRAMA†\cite{yuan2024drama} & 98.0 & 93.1 & 80.1 & 94.8 & 100 & 85.5 \\
DiffusionDrive†\cite{liao2025diffusiondrive} & 98.2 & 96.2 & 82.2 & 94.7 & 100 & 88.1 \\
\midrule
% 第二组方法
Hydra-MDP†\cite{li2024hydra} & 98.3 & 96.0 & 78.7 & 94.6 & 100 & 86.5 \\
Hydra-MDP‡\cite{li2024hydra} & 98.4 & 97.8 & 86.5 & 93.9 & 100 & 90.3 \\
Hydra-MDP§\cite{li2024hydra} & 98.4 & 97.7 & 85.0 & 94.5 & 100 & 89.9 \\
\midrule
% 第三组方法
DriveSuprim†\cite{yao2025drivesuprim} & 97.8 & 97.3 & 86.7 & 93.6 & 100 & 89.9 \\
DriveSuprim‡\cite{yao2025drivesuprim} & 98.0 & 98.2 & 90.0 & 94.2 & 100 & 92.1 \\
DriveSuprim§\cite{yao2025drivesuprim} & 98.6 & \textbf{98.6} & 91.3 & 95.5 & 100 & 93.5 \\
\midrule
TrajHF†(GRPO)\cite{li2025finetuning} & 98.6 & \textbf{98.6} & 91.3 & 95.5 & 100 & 94.0 \\

\midrule
% 第四组方法
EvaDrive†(\textcolor{conservative}{Cons})(Ours)\cellcolor{waymolgray}   & \textbf{98.9}\cellcolor{waymolgray}  & \textbf{98.6}\cellcolor{waymolgray}  & 90.5\cellcolor{waymolgray}  & \textbf{96.4}\cellcolor{waymolgray}  & \textbf{100}\cellcolor{waymolgray}  & 93.5\cellcolor{waymolgray}  \\
EvaDrive†(\textcolor{aggressive}{Aggres.})(Ours)\cellcolor{waymolgray}   & \textbf{98.9}\cellcolor{waymolgray}  & \underline{98.5}\cellcolor{waymolgray}  & \textbf{93.9}\cellcolor{waymolgray}  & \underline{96.2}\cellcolor{waymolgray}  & \textbf{100}\cellcolor{waymolgray}  & \textbf{94.9}\cellcolor{waymolgray}  \\
\bottomrule
\end{tabular}

\caption{Open-loop Results with Closed-loop Metrics on NAVSIMv1 Benchmark. Rows with \textcolor{conservative}{Cons}, and \textcolor{aggressive}{Aggres.} denote EvoDrive under conservative and aggressive driving styles respectively.†: ResNet34; ‡:V2-99; §:ViT-L}
\label{tab:navsim_eval}

\small  % 脚注说明保持小字体（符合规范）
\end{table}

\begin{table}[t!]
\centering
% 设置9号字体（字体大小9pt，行间距10.8pt）
\fontsize{9pt}{10.8pt}\selectfont
% 调整行间距和列间距
\renewcommand{\arraystretch}{1.3}  % 行间距系数
\setlength{\tabcolsep}{2pt}        % 列间距（可按需微调）

% 表格主体（无缩放命令，保持字体可控）
\begin{tabular}{c *{4}{c}}  % 1列方法名 + 4列指标，均居中对齐
\toprule  % 顶部粗横线
\textbf{Method} & \textbf{SR} & \textbf{DS} & \textbf{Effi.} & \textbf{Comf.}  \\
\midrule  % 中间横线
% 对比方法数据
AD-MLP\cite{zhai2023ADMLP} & 0.00 & 18.05 & 48.45 & 22.63 \\
UniAD\cite{hu2023planning} & 16.36 & 45.81 & 129.21 & 43.58 \\
VAD\cite{jiang2023vad} & 15.00 & 42.35 & \textbf{157.94} & \textbf{46.01} \\
TCP\cite{wu2022trajectoryguided} & 30.00 & 59.90 & 76.54 & 18.08 \\
ThinkTwice\cite{jia2023thinktwice} & 31.23 & 62.44 & 69.33 & 16.22 \\
DriveAdapter\cite{jia2023driveadapter} & 33.08 & 64.22 & 70.22 & 16.01 \\
DriveTransformer\cite{jia2025drivetransformer} & 35.01 & \textbf{65.02} & 100.64 & 20.78 \\

\midrule  % 分组横线
% 自研方法数据（带背景色）
EvaDrive(\textcolor{aggressive}{Aggres.})(Ours)\cellcolor{waymolgray} & 
\textbf{40.45}\cellcolor{waymolgray} & 
64.96\cellcolor{waymolgray} & 
130.21\cellcolor{waymolgray} & 
18.80\cellcolor{waymolgray} \\
\bottomrule  % 底部粗横线
\end{tabular}

% 表格标题（含指标缩写解释）
\caption{Closed-loop Results on Bench2Drive Benchmark. 
Abbreviations: SR = Success Rate, DS = Driving Score, Effi. = Efficiency, Comf. = Comfortness.}
\label{tab:b2d_eval}  % 表格标签

\end{table}

\begin{table}[ht]
\centering
\fontsize{9pt}{10.8pt}\selectfont
\renewcommand{\arraystretch}{1.0}
\setlength{\tabcolsep}{2pt} % 调整列间距以适配完整表头

\begin{tabular}{lcccccc}
    \toprule
    & \multicolumn{5}{c}{\textbf{Modules}} \\ 
    \cmidrule(lr){2-6}
    \multirow{2}{*}{\textbf{Variant}} & 
    \multirow{2}{*}{\shortstack{2-Stage\\Generator}} & 
    \multirow{2}{*}{\shortstack{Multi-Object.\\Reward}} & 
    \multirow{2}{*}{\shortstack{Multi-\\Turn}} & 
    \multirow{2}{*}{\shortstack{Pareto\\Guidance}} & 
    \multirow{2}{*}{\textbf{APO}} & 
    \multirow{2}{*}{\textbf{PDMS}} \\
    & & & & & & \\ 
    \midrule
    (S0)        & \ding{55} & \ding{55} & \ding{55} & \ding{55} & \ding{55} & 83.1 \\
    (S1)        & \ding{51}\cellcolor{waymollblue} & \ding{55} & \ding{55} & \ding{55} & \ding{55} & 88.1 \\
    (S2)        & \ding{51}\cellcolor{waymollblue} & \ding{51}\cellcolor{waymollblue} & \ding{55} & \ding{55} & \ding{55} & 91.7 \\
    (S3)        & \ding{51}\cellcolor{waymollblue} & \ding{51}\cellcolor{waymollblue} & \ding{51}\cellcolor{waymollblue} & \ding{55} & \ding{55} & 93.7  \\
    (S4)        & \ding{51}\cellcolor{waymollblue} & \ding{51}\cellcolor{waymollblue} & \ding{51}\cellcolor{waymollblue} & \ding{51}\cellcolor{waymollblue} & \ding{55} & 94.2\\
    (S5)        & \ding{51}\cellcolor{waymollblue} & \ding{51}\cellcolor{waymollblue} & \ding{51}\cellcolor{waymollblue} & \ding{51}\cellcolor{waymollblue} & \ding{51}\cellcolor{waymollblue} &  \textbf{94.9}\\
    \bottomrule
  \end{tabular}

\caption{Roadmap-based ablation study of EvaDrive. Modules are incrementally added.}
\label{tab:ablation_roadmap}
\end{table}

\section{Experiments}

Due to space constraints, we present the \textbf{three core experiments}; more experimental tables can be found in the appendix.
%\subsection{Different Driving Styles}

\subsection{Implementation Details}
Experiments are conducted on the NAVSIM~\cite{dauner2024navsim}, including open-loop evaluations on NAVSIM v1 (PDMS metric) and v2 (EPDMS metric) using real-world data, as well as closed-loop simulation on Bench2Drive via CARLA~\cite{Dosovitskiy17}. Our model uses a ResNet34 as backbone, MLP-based ego state encoders and reward models, and a 3-camera setup. Training is performed on 4 NVIDIA H20 GPUs with Adam\cite{kingma2014adam} optimizer (batch size 8 per GPU, learning rate \(7.5 \times 10^{-5}\)), alternating 5-epoch cycles between generator and critic for 30 epochs. Dataset and metric details are in the appendix.

\subsection{Quantitative and Qualitative Comparison}
\subsubsection{Quantitative Analysis.} Table~\ref{tab:navsim_eval} shows that EvaDrive achieves state-of-the-art results on NAVSIM v1, reaching 94.9 PDMS—surpassing DiffusionDrive\cite{liao2025diffusiondrive} by 6.8, DriveSuprim\cite{yao2025drivesuprim} by 5.0, and TrajHF by 0.9 without relying on external preference data. By tuning the weight vector $\mathbf{w} = \{w_i\}_{i=1}^N$, we implement diverse driving styles, including conservative (favoring safety, low EP) and aggressive (higher EP, proactive behaviors). On Bench2Drive in CARLA, EvaDrive attains a Driving Score of 64.96 (Table~\ref{tab:b2d_eval}), validating its closed-loop capability.
\subsubsection{Qualitative Analysis.}
Figure~\ref{fig:vis1} compares trajectories from GT, DiffusionDrive\cite{liao2025diffusiondrive}, and EvaDrive. Unlike the fixed behaviors of baselines, EvaDrive exhibits smooth transitions between conservative and aggressive modes via adjusting \(w\), without manual rules. Figure~\ref{fig:vis2} further illustrates the evolution of Pareto front size and distribution over iterations, highlighting the ability to explore diverse, high-quality solutions in the multi-objective space.

\subsection{Roadmap Ablation}

To systematically evaluate key components in \textit{EvaDrive}, we conduct a roadmap-style ablation study. Starting from a basic imitation learning baseline (\textbf{S0}), we incrementally introduce the two-stage generator, multi-objective reward model, multi-turn optimization, Pareto front guidance, and adversarial preference optimization. Table~\ref{tab:ablation_roadmap} summarizes each module's impact on planning performance over the NAVSIM v1 benchmark. 
Additional ablation results and analyses are provided in the appendix.

\textbf{S1}: Introducing the \textit{two-stage generator} significantly improves trajectory smoothness and spatial diversity (+5.0 PDMS). The autoregressive stage captures temporal intent, while the diffusion-based refiner enhances spatial flexibility.

\textbf{S2}: Building upon S1, we incorporate a multi-objective reward model encoding diverse criteria (e.g., collision, acceleration, ego-progress) as explicit signals. This enables fine-grained trade-offs between comfort and efficiency, boosting PDMS to 91.7.

\textbf{S3}: Adding \textit{multi-turn optimization} enables iterative refinement via historical feedback, enhancing robustness through continuous correction and reducing local optima (+2.0 PDMS vs. S2, reaching 93.7).

% \textbf{S4}: Enabling \textit{Pareto front guidance} preserves optimal multi-objective solutions, enhancing diversity while maintaining quality. This fosters adaptive behaviors, achieving 94.2 PDMS.

% \textbf{S5}: Introducing \textit{adversarial preference optimization} further improves policy quality. By training reward-aware discriminators and refining the actor via adversarial feedback, we discover superior trajectories, yielding the highest PDMS of \textbf{94.9}.
\textbf{S4}: \textit{Pareto front guidance} preserves optimal trade-offs across objectives, promoting diverse yet high-quality solutions and achieving a PDMS of 94.2.

\textbf{S5}: \textit{Adversarial preference optimization} enhances policy quality by training reward-aware discriminators to refine the actor via feedback, yielding the highest PDMS of \textbf{94.9}.

%\input{sec/exps/ablation_simple}

% \section{Conclusion}
% This paper introduces EvaDrive, a novel autonomous driving trajectory planner that unifies generation and evaluation via adversarial co-evolution. Framed as multi-objective RL (inspired by human decision-making), it uses multi-round Pareto optimization with a structured non-scalar reward model to preserve real-world driving diversity, avoiding flaws of scalar rewards or noisy annotations. Integrating a hierarchical generator, diffusion refinement, and multi-objective critic, it enables fine control over trajectory semantics, diversity, and safety. 
% % Experiments confirm its effectiveness in complex environments, offering a principled framework to advance closed-loop planning in autonomous systems.
% Experiments on NAVSIM and Bench2Drive benchmarks validate the effectiveness of EvaDrive in complex dynamic environments. It achieves state-of-the-art results: 94.9 PDMS on NAVSIM v1 and 64.96 Driving Score on Bench2Drive. These results highlight EvaDrive as the first adversarial closed-loop planning framework capable of scalarization-free, preference-aware trajectory optimization in both open-loop and closed-loop settings.
\section{Conclusion}
This paper presents \textbf{EvaDrive}, a multi-objective RL planner unifying generation and evaluation via adversarial co-evolution. Drawing on human decision-making, it uses multi-round Pareto optimization with a structured non-scalar reward model to maintain trajectory diversity while avoiding scalarization bias and annotation noise. Combining a hierarchical generator, diffusion-based refinement, and critic, EvaDrive enables precise control over trajectory semantics, diversity, and safety. Evaluated on NAVSIM and Bench2Drive, it achieves 94.9 PDMS and 64.96 Driving Score, positioning EvaDrive as the first closed-loop planner for scalarization-free, preference-aware trajectory optimization in both open- and closed-loop settings.

%\newpage

\section*{Appendix}
\vspace{1em}
\hrule
\vspace{1em}

\section{Further Related Works}
\subsection{E2E AD}
Imitation learning (IL) has been widely adopted for E2E trajectory planning in autonomous driving, where raw sensor inputs are mapped to expert demonstrations~\cite{shao2023safety, qian2024fasionad, qian2025fasionad++}. UniAD~\cite{hu2023planning}, Transfuser~\cite{chitta2022transfuser} leverage Bird-Eye-View (BEV) representations and introduce unified multi-task and safety-aware architectures to build planning-centric frameworks. VAD~\cite{jiang2023vad} further improve this design by employing query-based representations and constructing a trajectory vocabulary, effectively transforming the problem from regression into a discrete classification task.

To model the interactions between the ego vehicle and other traffic participants, GenAD~\cite{zheng2024genad} and PPAD~\cite{chen2024ppad} utilize instance-level visual features, while GraphAD~\cite{zhang2024graphad} represents these features as nodes in a graph structure. SparseDrive~\cite{sun2024sparsedrive} propose a fully sparse architecture, achieving greater efficiency and superior planning performance. However, most of these methods typically rely on constructing heavy BEV features, which prevents downstream tasks from learning directly from raw sensor data.

To address this issue, PARA-Drive~\cite{weng2024drive} and DriveTransformer~\cite{jia2025drivetransformer} introduce a parallel pipeline, enabling a more unified and scalable framework for end-to-end driving systems. Additionally, methods such as AD-MLP~\cite{zhai2023ADMLP} adopt a simple network architecture and are tested on nuScenes~\cite{caesar2020nuscenes}, relying solely on the ego vehicle's status information. These approaches highlight the potential over-reliance on ego status in simpler driving scenarios while also revealing limitations in the effective use of perception data.
Recent benchmarks like NavSim~\cite{dauner2024navsim} advance E2E planning with multi-agent scenarios, while frameworks including Hydra-MDP~\cite{li2024hydra}, GTRS~\cite{li2025generalized} and iPad~\cite{guo2025ipad} employ scorer-based trajectory synthesis integrating environmental interactions. However, these universally lack generator-scorer feedback loops. Contrastingly, our human decision-inspired solution integrates trajectory generation and evaluation within an iterative optimization cycle.

\subsection{RL in generative task(multi-object and multi-turn)}
%主要讲多轮和多目标这两个, AI找的文献，幻觉感觉很大的
% Chen et al. (2023). Memory-Enhanced Dialogue Policy for Multi-Turn Interaction. ACL.
% Feng et al. (2020). Curriculum Learning for Dialogue Policy Optimization. NeurIPS.
% Guo et al. (2021). Actor-Critic with Memory Networks for Long-Horizon Dialogues. EMNLP.
% He et al. (2022). Inverse RL for Preference-Aware Dialogue Systems. ICML.
% Kostrikov et al. (2021). Offline Reinforcement Learning for Dialogue. ICLR.
% Li et al. (2020). Hierarchical RL for Multi-Objective Dialogues. AAAI.
% Ouyang et al. (2022). Training Language Models to Follow Instructions with RL. NeurIPS.
% Wang et al. (2019). Weighted Sum Methods in Dialogue Policy Optimization. ACL.
% Zhang et al. (2021). Pareto-Based Multi-Objective Dialogue Generation. EACL.
% Zhao et al. (2022). Hierarchical Dialogue Policy Optimization. COLING.
RL has emerged as a critical paradigm for optimizing generative models in complex, goal-driven scenarios, particularly in multi-objective and multi-turn dialogue systems\cite{ouyang2022training, zhou2025safemvdrive, shani2024multi}. These tasks require balancing competing objectives (e.g., fluency, coherence, task success) and maintaining long-term context consistency across multiple interaction turns.

% Key challenges include reward engineering for non-differentiable objectives, exploration-exploitation trade-offs in sparse reward settings, and generalization across diverse user behaviors. Emerging directions focus on integrating RL with large language models (LLMs) for reward modeling\cite{ouyang2022training} and leveraging offline RL to reduce real-time interaction costs\cite{gupta2023offline}. Hybrid approaches combining RL with pre-trained dialogue systems (e.g., DialoGPT\cite{yang2021ubar}) show promise in achieving both task efficiency and natural language fluency. In autonomous driving, reinforcement learning (RL) methods like GRPO~\cite{shao2024deepseekmath} and DPO~\cite{rafailov2023direct} are typically deployed during post-training to align trajectories with human preferences. By contrast, our framework employs \textit{multi-round multi-objective RL} with APO method~\cite{cheng2023adversarial} for trajectory optimization.
Key challenges include non-differentiable reward engineering, sparse-reward exploration-exploitation tradeoffs, and cross-user generalization. Emerging solutions integrate LLMs for reward modeling~\cite{ouyang2022training} and offline RL to reduce interaction costs~\cite{gupta2023offline}, while hybrid methods with pretrained dialogue systems~\cite{yang2021ubar} balance task efficiency with natural fluency. In autonomous driving, post-training alignment methods like GRPO~\cite{shao2024deepseekmath} and DPO~\cite{rafailov2023direct} optimize human preference matching. Contrastingly, our framework pioneers multi-round multi-objective RL with Adversarial Policy Optimization (APO)~\cite{cheng2023adversarial} for trajectory optimization.

% \iffalse
\subsection{GAN in generative task}
% Chen et al. (2023). DiffusionGAN for Robotic Manipulation. ICRA.
% Isola et al. (2017). Image-to-Image Translation with Conditional Adversarial Networks. CVPR.
% Karras et al. (2018). Progressive Growing of GANs for Improved Quality and Stability. ICLR.
% Luo et al. (2020). Domain Adaptation for Nighttime Perception via CycleGAN. NeurIPS.
% Pumarola et al. (2020). D-Plane: 3D-Aware Image Synthesis. ECCV.
% Wang et al. (2021).PhysGAN: Generating Physical-World-Resilient Adversarial Examples for Autonomous Driving.
% Zhu et al. (2017). Unpaired Image-to-Image Translation Using Cycle-Consistent Adversarial Networks. ICCV.
Generative Adversarial Networks (GANs) have emerged as a cornerstone of modern generative modeling, enabling the synthesis of high-quality, realistic data across modalities\cite{goodfellow2014generative}. Their adversarial training framework, comprising a generator and discriminator in a minimax game, has driven breakthroughs in image generation\cite{karrast2018progressive}, text-to-image synthesis\cite{kang2023scaling}, and domain adaptation\cite{zhu2017unpaired}. 
% GANs have achieved unprecedented realism in image generation, particularly through architectural innovations. The introduction of deep convolutional architectures (Radford et al., 2015) and progressive growth strategies (Karras et al., 2018) enabled high-resolution outputs, culminating in state-of-the-art models like StyleGAN (Karras et al., 2019) and StyleGAN2 (Karras et al., 2020). These models decouple semantic attributes from stochastic variations, allowing fine-grained control over features such as facial expressions or object textures. For instance, StyleGAN2’s latent space manipulation has become a standard for editing real-world images (Richardson et al., 2021). Meanwhile, conditional GANs (cGANs) like Pix2Pix (Isola et al., 2017) and Pix2PixHD (Wang et al., 2018) bridge label-to-image translation, underpinning applications from medical imaging to scene rendering.

In autonomous driving, GANs address critical challenges in data scarcity, sensor simulation, and domain adaptation. First, they generate synthetic sensor data (e.g., RGB-D) to augment training datasets. CycleGAN\cite{zhu2017unpaired} has been adapted to translate daytime images to nighttime abstractscenes for robust perception under varying lighting conditions\cite{yang2020surfelgan}. SocialGAN~\cite{gupta2018social} integrates adversarial training for pedestrain motion forecasting. Physics-informed GANs like PhySGAN~\cite{kong2020physgan} enforce physical consistency in simulated sensor outputs by incorporating collision dynamics, reducing the simulation-to-reality gap. This paper integrates GANs into a reinforcement learning framework for autonomous driving, addressing critical gaps in trajectory planning research.
% \fi
% The task of trajectory generation has developed rapidly along with the advancement of autonomous driving

\section{Further Loss Function Detail}
\subsection{Training Objectives}
To jointly optimize the \textbf{generator (Actor)} and the \textbf{multi-objective reward model (Critic)}, we formulate a structured loss design. We emphasize a \textbf{multi-objective design philosophy}: each reward dimension reflects a distinct aspect of trajectory quality (e.g., safety, comfort, efficiency). To accommodate this, we introduce a \textbf{normalized weight vector} $\{w_i\}_{i=1}^N$, with $w_i \geq 0$ and $\sum_i w_i = 1$, enabling adaptive emphasis on different trajectory qualities. These weights can be either manually defined or learned during training, as discussed later.

\subsection{Actor Loss ($L_{\text{actor}}$)}
The Actor $\pi_\theta$ aims to generate high-quality trajectories that are not only feasible but also align with desired behavioral preferences. The loss consists of three components:

\paragraph{(1) Imitation Loss.}
The \textbf{imitation loss} encourages the Actor $\pi_\theta$ to mimic expert demonstrations. This is implemented as a supervised loss over state-action pairs, specifically using the \textbf{Binary Cross-Entropy (BCE) loss}, formulated as:
\begin{equation}
L_{\text{imitation}} = \mathbb{E}_{(s,a) \sim \pi_{\mathrm{expert}}} \left[ \mathrm{BCE}(\pi_\theta(a|s), a) \right].
\end{equation}
Here, $s$ denotes the \textbf{state}, $a$ denotes the \textbf{action}, $\pi_\theta(a|s)$ is the Actor's predicted action distribution given state $s$, and $\pi_{\mathrm{expert}}$ represents the \textbf{expert policy}.

\paragraph{(2) Diversity Loss.}
To mitigate mode collapse and promote exploration, we add a \textbf{diversity loss} that penalizes semantic redundancy among generated trajectories within the same batch. We encourage the policy to cover a broader distribution by \textbf{minimizing mutual information}:
\begin{equation}
L_{\text{diversity}} = \mathbb{E}_{\substack{\tau_i, \tau_j \sim \pi_\theta \\ i \neq j}} \left[ \mathrm{MI}(\tau_i, \tau_j) \right],
\end{equation}
where $\tau_i$ and $\tau_j$ are \textbf{distinct trajectories} generated by the Actor $\pi_\theta$, and $\mathrm{MI}(\cdot, \cdot)$ denotes \textbf{mutual information}, used to measure the redundancy between encodings of two trajectories.

\paragraph{(3) Adversarial Preference Loss.}
This is a key innovation: we formulate trajectory generation as a \textbf{multi-objective reward maximization problem}, using a learned reward model $r_\phi^{(i)}(s,a)$ for each dimension. The Actor seeks to generate trajectories that maximize cumulative reward across all dimensions:
\begin{equation}
L_{\text{adv, actor}} = - \mathbb{E}_{(s,a) \sim \pi_\theta} \left[ \sum_{i=1}^N w_i r_\phi^{(i)}(s,a) \right] + \lambda \cdot \mathrm{KL}(\pi_\theta \| \pi_{\mathrm{ref}}).
\end{equation}
Here, $r_\phi^{(i)}(s,a)$ is the Critic model's predicted reward for the state-action pair $(s,a)$ on the \textbf{$i$-th reward dimension}, and $N$ is the \textbf{total number of reward dimensions}. $\lambda$ is the \textbf{KL regularization coefficient}, and $\mathrm{KL}(\pi_\theta \| \pi_{\mathrm{ref}})$ is the \textbf{KL divergence} between the Actor's policy $\pi_\theta$ and a \textbf{reference policy} $\pi_{\mathrm{ref}}$, used to stabilize training by constraining deviations from the reference policy.

\textbf{Flexible Objective Control:} We attempted to parameterize the weights $\{w_i\}$ and optimize them through training, but the results were unsatisfactory. The model tended to directly adjust the weights in directions that made loss reduction easier, ultimately leading to mode collapse. Therefore, we switched to manually designing several different weight configurations; relevant experimental data can be found in Table 7. 

\paragraph{Final Actor Objective:}
\begin{equation}
L_{\text{actor}} = L_{\text{imitation}} + L_{\text{diversity}} + L_{\text{adv, actor}}.
\end{equation}

\subsection{Critic Loss ($L_{\text{critic}}$)}
The Critic, a \textbf{multi-objective reward model} $r_\phi = \{r_\phi^{(i)}\}_{i=1}^N$, is trained to evaluate the quality of trajectories along distinct reward dimensions. Its loss includes:

\paragraph{(1) Reward Supervision Loss.}
When ground-truth or human-labeled reward signals are available, we use a supervised regression loss to train each reward head. This term explicitly employs the \textbf{Binary Cross-Entropy (BCE) loss}:
\begin{equation}
L_{\text{reward}} = \mathbb{E}_{(s,a)} \left[ \sum_{i=1}^N \mathrm{BCE} \left(r_\phi^{(i)}(s,a), r_{\text{gt}}^{(i)}(s,a) \right) \right],
\end{equation}
where $r_\phi^{(i)}(s,a)$ is the Critic model's predicted reward on the \textbf{$i$-th dimension}, and $r_{\text{gt}}^{(i)}(s,a)$ is the \textbf{ground-truth reward signal for the $i$-th dimension}. This term anchors the Critic's predictions in observable feedback, consistent with the "BCE loss" paradigm in reinforcement learning.

\paragraph{(2) Adversarial Alignment Loss.}
In the adversarial loop, the Critic is optimized to distinguish expert trajectories from generated ones in the multi-dimensional reward space:
\begin{align}
L_{\text{adv, critic}} = & \ \mathbb{E}_{(s,a) \sim \pi_\theta} \left[ - \sum_{i=1}^N w_i r_\phi^{(i)}(s,a) \right] \\
& + \mathbb{E}_{(s,a) \sim \pi_{\mathrm{expert}}} \left[ \sum_{i=1}^N w_i r_\phi^{(i)}(s,a) \right] \nonumber \\
& + \beta \cdot \mathrm{KL}(P_{\mathrm{expert}} \| Q_\phi),
\end{align}
where $\beta$ is the \textbf{KL regularization coefficient}, $P_{\mathrm{expert}}$ is the \textbf{reward distribution of expert trajectories}, and $Q_\phi$ denotes the \textbf{Critic-induced distribution over preferences}. The KL regularization prevents overfitting to the expert dataset.

\paragraph{Final Critic Objective:}
\begin{equation}
L_{\text{critic}} = L_{\text{reward}} + L_{\text{adv, critic}}.
\end{equation}

This collaborative training scheme establishes a dynamic interplay between Actor and Critic: the Actor generates trajectories to maximize multi-dimensional rewards, while the Critic continually refines its judgment to distinguish superior behaviors. This adversarial preference optimization drives the model to explore high-reward, diverse, and controllable behaviors across iterations. The overall training objective is defined as a combination of actor and critic losses:
\begin{equation}
L = L_{\text{actor}} + L_{\text{critic}}.
\end{equation}

\begin{algorithm}[h]
\caption{Differentiable Fast Non-dominated Sorting with Gumbel-Softmax Sampling}
\label{alg:fast-ns-gumbel}
\SetAlgoVlined
\SetCommentSty{gray}
\KwIn{Candidate trajectories $\mathcal{A} = \{a_i\}_{i=1}^{64}$, reward vectors $\{\mathbf{r}(a_i)\}_{i=1}^{64}$}
\KwOut{Sampled trajectories $\{\tilde{a}^{(m)}\}_{m=1}^M$ for differentiable guidance}

\textbf{1. Fast non-dominated sorting} \\
Initialize $n_i \leftarrow 0$, $S_i \leftarrow \emptyset$, $\mathcal{F}_1 \leftarrow \emptyset$\; 
\ForEach{$a_i \in \mathcal{A}$}{
  \ForEach{$a_j \in \mathcal{A}$}{
    \lIf{$\mathbf{r}(a_i) \succ \mathbf{r}(a_j)$}{ $S_i \leftarrow S_i \cup \{a_j\}$ }
    \lElseIf{$\mathbf{r}(a_j) \succ \mathbf{r}(a_i)$}{ $n_i \leftarrow n_i + 1$ }
  }
  \lIf{$n_i = 0$}{ $r_i \leftarrow 1$, $\mathcal{F}_1 \leftarrow \mathcal{F}_1 \cup \{a_i\}$ }
}
$k \leftarrow 1$\;
\While{$\mathcal{F}_k \neq \emptyset$}{
  $\mathcal{Q} \leftarrow \emptyset$\;
  \ForEach{$a_i \in \mathcal{F}_k$}{
    \ForEach{$a_j \in S_i$}{
      $n_j \leftarrow n_j - 1$\;
      \lIf{$n_j = 0$}{ $r_j \leftarrow k+1$, $\mathcal{Q} \leftarrow \mathcal{Q} \cup \{a_j\}$ }
    }
  }
  $k \leftarrow k + 1$, $\mathcal{F}_k \leftarrow \mathcal{Q}$\;
}

\textbf{2. Extract Pareto front} $\mathcal{P} \leftarrow \mathcal{F}_1$\;

\textbf{3. Compute crowding distance} $\{d(a)\}_{a \in \mathcal{P}} \leftarrow \text{CrowdingDistance}(\mathcal{P})$\;

\textbf{4. Gumbel-Softmax sampling from $\mathcal{P}$} \\
Compute logits $\ell(a) \propto d(a)$ for $a \in \mathcal{P}$\;
\For{$m = 1$ \KwTo $M$}{
  $\tilde{a}^{(m)} \sim \text{GumbelSoftmaxSample}(\{\ell(a)\}_{a \in \mathcal{P}})$\;
}
\Return $\{\tilde{a}^{(m)}\}_{m=1}^M$\;
\end{algorithm}

\section{Further Visualization Results}
We present additional visualization results. Figure 5 shows comparative visualizations of our method, ground truth (GT), and DiffusionDrive across more road scenarios, demonstrating the consistency of our approach in diverse environmental conditions. Figure 6 visualizes the trajectory candidates generated by our model and the constructed Pareto frontier.These visualizations further validate the effectiveness and rationality of our method in practical driving scenarios.
\begin{figure*}[ht]
\centering
\includegraphics[width=0.98\linewidth]{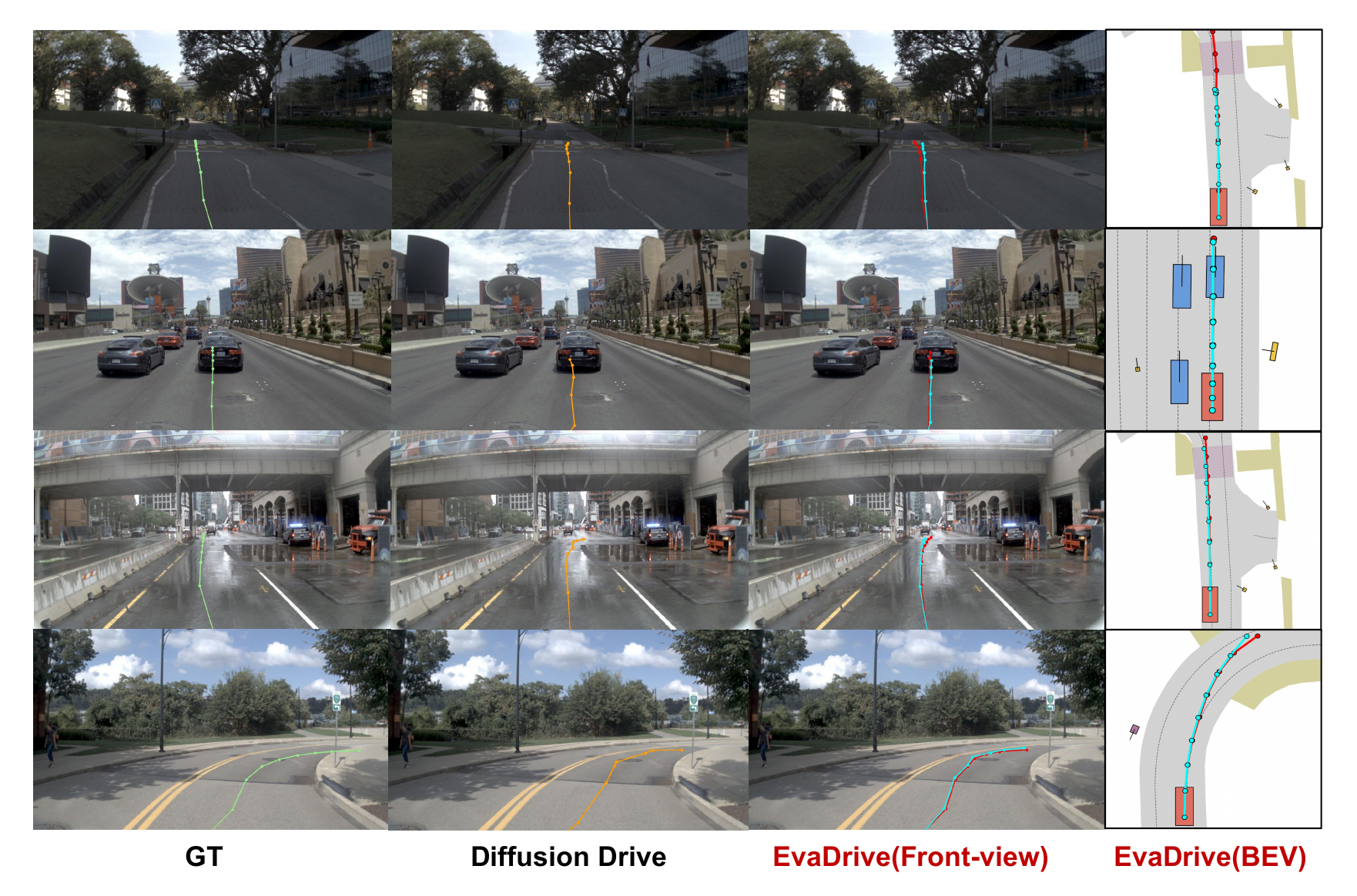}
\caption{Comparative visualizations of ground truth, DiffusionDrive, and our method trajectories across diverse road scenarios (urban intersections, highway merges) on front camera views.}
\label{fig:traj_comparison}
\end{figure*}
\begin{figure*}[ht]
\centering
\includegraphics[width=0.98\linewidth]{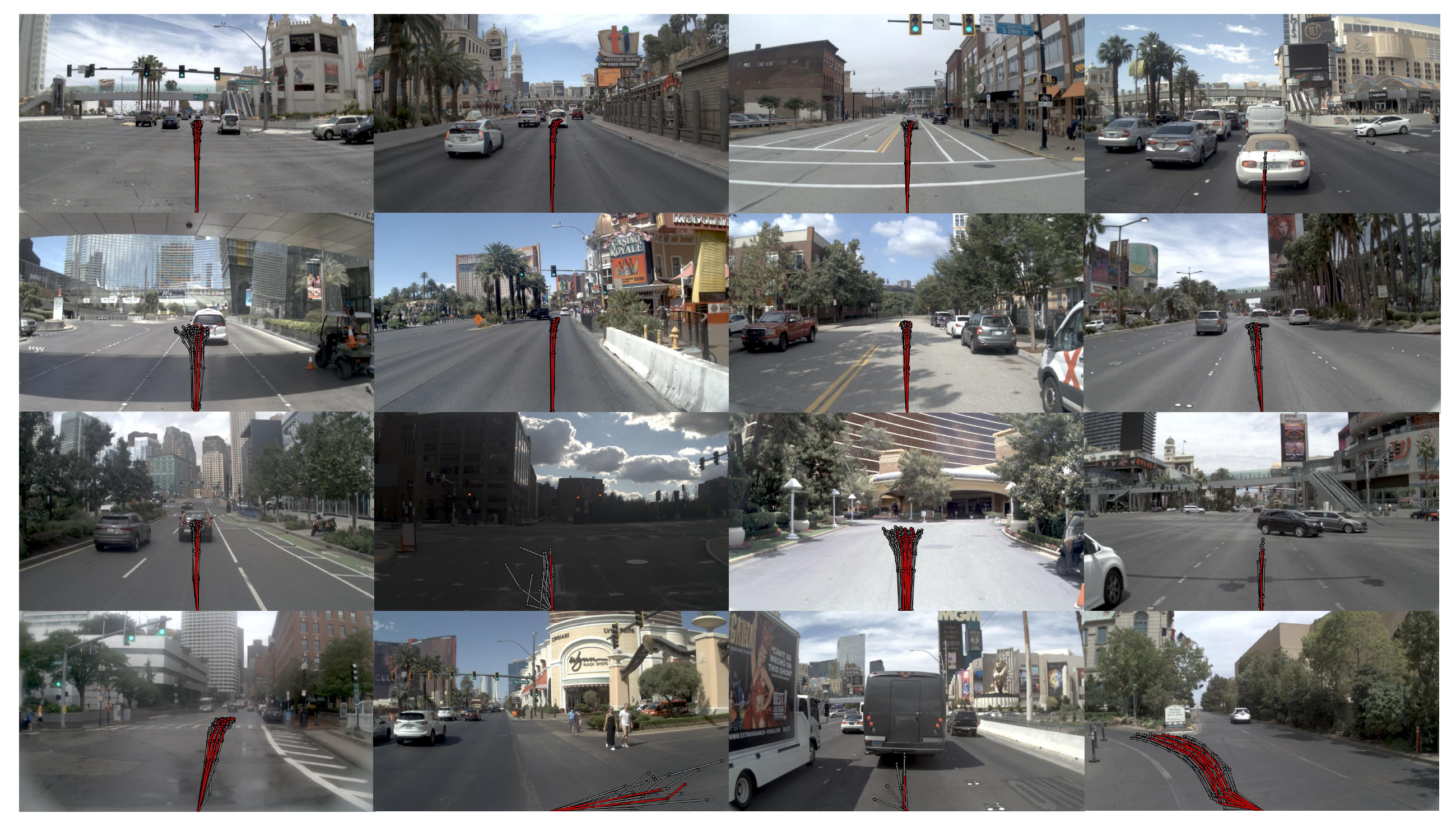}
\caption{Visualization of model-generated trajectory candidates (gray lines) and the Pareto frontier in red, highlighting optimal safety-efficiency-comfort trade-offs.}
\label{fig:pareto_frontier}
\end{figure*}
\begin{table*}[htbp]
\centering

{\small % 使用小字体，可根据需要换成\footnotesize或\tiny等
\begin{tabular}{llcccccccccc}
\toprule
\textbf{Method} & \textbf{Backbone} & \textbf{NC}$\uparrow$ & \textbf{DAC}$\uparrow$ & \textbf{DDC}$\uparrow$ & \textbf{TL}$\uparrow$ & \textbf{EP}$\uparrow$ & \textbf{TTC}$\uparrow$ & \textbf{LK}$\uparrow$ & \textbf{HC}$\uparrow$ & \textbf{EC}$\uparrow$ & \textbf{EPDMS}$\uparrow$ \\
\midrule
Human Agent & — & 100 & 100 & 99.8 & 100 & 87.4 & 100 & 100 & 98.1 & 90.1 & 90.3 \\
EgoStatusMLP & — & 93.1 & 77.9 & 92.7 & 99.6 & 86.0 & 91.5 & 89.4 & 98.3 & 85.4 & 64.0 \\
\midrule
Transfuser[7] & ResNet34 & 96.9 & 89.9 & 97.8 & 99.7 & 87.1 & 95.4 & 92.7 & 98.3 & 87.2 & 76.7 \\
\midrule
HydraMDP++[27] & ResNet34 & 97.2 & 97.5 & 99.4 & 99.6 & 83.1 & 96.5 & 94.4 & 98.2 & 70.9 & 81.4 \\
HydraMDP++[27] & V2-99 & 98.4 & 98.0 & 99.4 & 99.8 & 87.5 & 97.7 & 95.3 & 98.3 & 77.4 & 85.1 \\
HydraMDP++[27] & ViT-L & 98.5 & 98.5 & 99.5 & 99.7 & 87.4 & 97.9 & 95.8 & 98.2 & 75.7 & 85.6 \\
\midrule
DriveSuprim & ResNet34 & 97.5 & 96.5 & 99.4 & 99.6 & 88.4 & 96.6 & 95.5 & 98.3 & 77.0 & 83.1 \\ 
DriveSuprim & V2-99 & 97.8 & 97.9 & 99.5 & 99.9 & 90.6 & 97.1 & 96.6 & 98.3 & 77.9 & 86.0 \\ 
\midrule
EvoDrive(\textcolor{aggressive}{Aggres.})& ResNet34 & 98.8 & 98.5 & 98.9 & 99.8 & 96.6 & 98.4 & 94.3 & 97.8 & 55.9 & 86.3 \\
\bottomrule
\end{tabular}
}
\caption{Evaluation on NAVSIM v2. Results are grouped by methods.} 
\label{tab:navsim_eval_2} 
\end{table*}

\section{Further Experiment Settings}

\begin{table*}[htpb]
\centering
\renewcommand{\arraystretch}{1.1}
\fontsize{10pt}{12pt}\selectfont  % 设置字体大小为10pt
\begin{tabular}{lccc}
\toprule
\textbf{Method} & \textbf{Reward Type} & \textbf{Optimization Method} & \textbf{PDMS ↑} \\
\midrule
PPO (Scalarized) & Single-objective & PPO & 91.7 \\
TrajHF (GRPO-style) & Single-objective & GRPO & 94.0 \\
EvaDrive (w/o APO) & Multi-objective & PPO & 93.5 \\
\textbf{EvaDrive (Full)} & Multi-objective & \textbf{APO} & \textbf{94.9} \\
\bottomrule
\end{tabular}
\caption{Ablation on reward structures and optimization strategies.Performance comparisons under different conditions (NAVSIM Benchmark)}
\label{tab:ablation_pdms}
\end{table*}

% 第一个表格：EvaDrive vs. Iterations
\begin{table*}[htpb]
\centering
\renewcommand{\arraystretch}{1.1}  % 调整宽度避免过宽
\fontsize{10pt}{12pt}\selectfont  % 设置字体大小为10pt
\begin{tabular}{l *{4}{c}}
\toprule
\textbf{Iterations} & \textbf{PDMS} & \textbf{Time(ms)} & \textbf{Avg. Pareto Size}& \textbf{Avg.Crowding Distance} \\
\midrule
1 (Single)          & 93.1           & 45                  & -             & -            \\
2                   & 94.9           & 69                  & 8.5          & 12.3       \\
3                   & 94.8           & 104                 & 6.1           & 9.3       \\
4                   & 93.6           & 145                 & 3.2          & 5.5       \\
\bottomrule
\end{tabular}
\caption{EvaDrive vs. Iterations (Performance comparisons under different conditions)}
\label{tab:iterations_performance_simplified}
\end{table*}

% 第二个表格：Different Preference Weights

\begin{table*}[htpb]
\centering
\renewcommand{\arraystretch}{1.1}
\fontsize{10pt}{12pt}\selectfont
\begin{tabular}{>{\centering\arraybackslash}m{3.5cm}*{6}{>{\centering\arraybackslash}m{0.8cm}}}
\toprule
\textbf{Preference} & \textbf{NC} & \textbf{DAC} & \textbf{EP} & \textbf{TTC} & \textbf{C} & \textbf{PDMS} \\
\midrule
\rowcolor{gray!20} Weight Parameterization (Self-Learned) & 89.3 & 90.1 & 76.5 & 88.2 & 91.7 & 87.2 \\
Conservative (0.4:0.2:0.2:0.1:0.1) & 99.1 & 98.8 & 87.2 & 97.1 & 100 & 93.5 \\
Balanced (0.2:0.2:0.2:0.2:0.2) & 98.9 & 98.5 & 91.9 & 96.2 & 100 & 94.2 \\
Aggressive (0.1:0.4:0.1:0.2:0.2) & 98.7 & 97.9 & 93.3 & 95.8 & 100 & 94.9 \\
Comfort-focused (0.2:0.2:0.4:0.1:0.1) & 99.0 & 98.2 & 89.0 & 96.5 & 100 & 94.0 \\
\bottomrule
\end{tabular}
\caption{Performance Comparisons Under Different Preference Weights (Safety:Efficiency:Comfort:EgoProgress:CollisionAvoidance) on NAVSIM Benchmark. Subscores include No Collisions (NC), Drivable Area Compliance (DAC), Ego Progress (EP), Time-to-Collision (TTC), and Comfort (C).}
\label{tab:preference_weights}
\end{table*}

% 第三个表格：Ablation on reward structures and optimization strategies

% 调整表格间间距（进一步缩小）
% 第一个表格：Two-Stage Actor 消融研究
% 第一个表格：Two-Stage Actor 消融研究（数值调整）
\begin{table*}[htpb]
\centering
\renewcommand{\arraystretch}{1.1}
\fontsize{10pt}{12pt}\selectfont
\begin{tabular}{lcccccc}
\toprule
\textbf{Experimental Setup} & \textbf{NC} & \textbf{DAC} & \textbf{TTC} & \textbf{C} & \textbf{EP} & \textbf{PDMS} \\
\midrule
No stages (baseline) & 97.8 & 95.2 & 94.0 & 99.5 & 81.5 & 83.1 \\
Only Stage 1 & 98.0 & 95.6 & 94.2 & 99.6 & 81.8 & 85.6 \\
Only Stage 2 & 98.2 & 96.3 & 94.6 & 99.8 & 82.0 & 87.0 \\
Reversed order (Stage 2 → Stage 1) & 97.9 & 95.8 & 94.3 & 99.7 & 82.2 & 85.9 \\
Our design (Stage 1 → Stage 2) & 98.3 & 96.5 & 94.8 & 100 & 83.3 & 88.1 \\
\bottomrule
\end{tabular}
\caption{Ablation study on Two-Stage Actor Design (NAVSIM Benchmark). "Stage 1 → Stage 2" denotes our proposed sequential architecture, while "Stage 2 → Stage 1" represents reversed module order.}
\label{tab:ablation_two_stage}
\end{table*}

% 第二个表格：Anchor Number 消融研究（数值调整）
\begin{table*}[htpb]
\centering

\renewcommand{\arraystretch}{1.1}
\fontsize{10pt}{12pt}\selectfont  % 统一字体大小为10pt
\begin{tabular}{lcccccc}
\toprule
\textbf{Config} & \textbf{NC} & \textbf{DAC} & \textbf{TTC} & \textbf{C} & \textbf{EP} & \textbf{PDMS} \\
\midrule
16 & 97.2 & 93.0 & 92.5 & 99.0 & 89.3 & 90.5 \\
32 & 98.0 & 95.9 & 94.5 & 99.7 & 93.3 & 92.9 \\
64 & 98.6 & 96.3 & 95.0 & 100 & 93.2 & 94.85 \\
128 & 98.6 & 96.4 & 95.1 & 100 & 91.3 & 94.85 \\
\bottomrule
\end{tabular}
\caption{Ablation study on Anchor Number (NAVSIM Benchmark)}
\label{tab:ablation_anchor_number}
\end{table*}

\subsection{Implementation Details}
\paragraph{Dataset and Metrics}
We conduct experiments mainly on the NAVSIM [9] benchmark, which is a driving dataset containing challenging driving scenarios. There are two different evaluation metrics on NAVSIM, leading to NAVSIMv1 and NAVSIMv2. The evaluation metric of NAVSIMv1 is the \textbf{PDM Score (PDMS)}. Each predicted trajectory is sent to a simulator to collect different rule-based subscores, which are aggregated to get the final PDMS:
% Formula
\begin{equation}
\label{eq:pdms}
\text{PDMS} = \frac{\prod_{m \in SP} \text{score}_m \times \sum_{w \in SA} \text{weight}_w \times \text{score}_w}{\sum_{w \in SA} \text{weight}_w}
\end{equation}
% End Formula

\section{Further Experiment Settings}
where $SP$ and $SA$ denote the penalty subscore set and the weighted average subscore set. In NAVSIMv1, $SP$ comprises two subscores, including \textbf{no collisions (NC)} and \textbf{drivable area compliance (DAC)}, and $SA$ comprises \textbf{ego progress (EP)}, \textbf{time-to-collision (TTC)}, and \textbf{comfort (C)}. For NAVSIM, we use the official log-replay simulator with an LQR controller operating at 10 Hz over a 4-second horizon, with detailed sub-metric definitions as follows: 

\begin{itemize}
    \item \textbf{NoAt-Fault Collision (NC)}: Set to 0 if, at any simulation step, the proposal’s bounding box intersects with other road users (vehicles, pedestrians, or bicycles). Collisions not considered “at-fault” in non-reactive environments (e.g., ego vehicle stationary) are ignored; collisions with static objects apply a softer penalty of 0.5. 
    \item \textbf{Drivable Area Compliance (DAC)}: Set to 0 if any corner of the proposal state lies outside drivable area polygons at any step. 
    \item \textbf{Time-to-Collision (TTC)}: Initialized to 1, set to 0 if projected TTC (assuming constant velocity/heading) drops below 1 second within the 4-second horizon. 
    \item \textbf{Comfort (C)}: Set to 0 if motion exceeds thresholds: lateral acceleration \textgreater\ 4.89 m/s², longitudinal acceleration \textgreater\ 2.40 m/s², longitudinal deceleration \textgreater\ 4.05 m/s², absolute jerk \textgreater\ 8.37 m/s³, longitudinal jerk \textgreater\ 4.13 m/s³, yawrate \textgreater\ 0.95 rad/s, or yaw acceleration \textgreater\ 1.93 rad/s².
    \item \textbf{Ego Progress (EP)}: Measures progress along the route center, normalized by a safe upper bound from the PDM-Closed planner. The ratio is clipped to [0,1], with scores discarded if the upper bound \textgreater\ 5 meters or progress is negative.
\end{itemize}

NAVSIMv2 introduces 4 additional subscores: \textbf{driving direction compliance (DDC)} and \textbf{traffic light compliance (TLC)} are added to the penalty subscore set (SP), while \textbf{lanekeeping (LK)} and \textbf{extended comfort (EC)} are included in the weighted average subscore set (SA). Additionally, the original comfort subscore is revised and renamed as \textbf{history comfort (HC)}. Aggregating all these subscores yields the \textbf{EPDMS} metric specific to NAVSIMv2. Table 4 presents the performance of our method on the NAVSIMv2 benchmark.

We also evaluate on the Bench2Drive benchmark, utilizing Carla simulator with a perfect controller operating at 2 Hz over a 3-second horizon. Its evaluation sub-metrics include: 
\begin{itemize}
    \item \textbf{NoCollision (NC)}: Set to 0 if the proposal’s bounding box intersects with any object (vehicles, bicycles, pedestrians, traffic signs, cones, or lights) at any step. 
    \item \textbf{Drivable Area Compliance (DAC)}: Set to 0 if any corner of the proposal state is off-road or all centers are off-route at any step. 
    \item \textbf{Time-to-Collision (TTC)}: Set to 0 if projected TTC drops below 1 second within the 3-second horizon. 
    \item \textbf{Comfort}: Set to 0 if acceleration or turning rate exceeds the expert trajectory’s maximum values. 
    \item \textbf{Ego Progress}: Defined as the ratio of ego progress along the expert trajectory (conditioned on collision-free and on-road status). If the ratio exceeds 1, its reciprocal is used.
\end{itemize}

\paragraph{Model Details}

Our method employs a ResNet34 backbone as the image encoder $\text{Enc}_i$. Input images are resized to a resolution of $2048 \times 512$. We use a two-layer MLP (Multi-Layer Perceptron) to encode the ego-vehicle state and the high-scoring trajectories returned by the critic, with the encoded results used for anchor initialization. Another two-layer MLP is adopted as the reward model, followed by a linear layer to predict all metrics.  

The number of candidate trajectories per round is set to 64, and the Pareto frontier is constructed using the \textbf{Fast Non-Dominated Sorting algorithm} (see Algorithm 2 for details on frontier construction and differentiable injection). The process involves: first, ranking the 64 candidate trajectories into hierarchical levels based on multi-objective metrics to form solution sets with different dominance ranks; then selecting the highest-ranked non-dominated solution set as the current Pareto frontier, with diversity maintained via crowded distance calculation to prevent local concentration. After construction, Gumbel-Softmax sampling injects frontier solutions into the previous round’s trajectory candidates, ensuring full differentiability throughout training.  

For perception input, we adopt a three-camera FOV setup, with images sourced from the \texttt{l0} (left), \texttt{f} (front), and \texttt{r0} (right) cameras.

\paragraph{Training Details}
We train our model on 4 NVIDIA H20 GPUs using a \textbf{generative-adversarial approach}. We adopt the \textbf{Adam optimizer} for model training, with the batch size on a GPU set to 8, and the learning rate at $7.5 \times 10^{-5}$. Our training strategy involves \textbf{alternating optimization cycles} between the generator and the critic. Specifically, the generator is trained for 5 epochs, followed by the critic for 5 epochs. This 5-epoch cycle is repeated alternately, for a total of \textbf{30 training epochs}.
\section{Further Ablation Studies}

\subsection{Reward Structures and Optimization Strategies}

To better understand the individual contributions of reward modeling and optimization mechanisms in our framework, we conduct a comprehensive ablation study covering both single-objective and multi-objective reinforcement learning paradigms. We distinguish methods along two axes: the \textit{reward type} (i.e., whether the policy is trained using a scalar objective or a structured reward vector), and the \textit{optimization strategy} (e.g., vanilla policy gradient, preference-based optimization, or our adversarial policy optimization paradigm).

Table~\ref{tab:ablation_pdms} summarizes the results in terms of PDMS, a holistic driving quality metric. As seen, the baseline method \textbf{PPO (Scalarized)} adopts manually weighted scalar rewards and uses standard policy optimization. Despite being simple and widely used, it fails to balance safety and comfort, resulting in the lowest PDMS.

We also evaluate \textbf{TrajHF}, a GRPO-style method which learns a scalar reward via pairwise preference rankings. Although it improves slightly over scalarized PPO, its scalar reward structure fundamentally limits its ability to represent multi-dimensional planning objectives. Therefore, we categorize TrajHF as a \textit{single-objective} RL method, despite the use of preference-based supervision.

To isolate the effect of multi-objective rewards, we compare with \textbf{EvaDrive (w/o APO)}, which leverages structured reward vectors but optimizes the policy without adversarial co-evolution or Pareto-guided feedback. This variant already shows notable gains, indicating that vectorized reward modeling improves alignment with planning principles.

Finally, our full framework \textbf{EvaDrive (APO)} achieves the best performance, demonstrating the synergy between structured reward modeling, adversarial optimization, and multi-turn Pareto-guided feedback. These results highlight that each component—multi-objective formulation, co-evolutionary training, and Pareto supervision—plays a complementary and indispensable role.

\subsection{Number of Multi-Round Iterations}
Table 6 investigates the impact of iteration counts on planning performance. Experimental results show that our method can achieve satisfactory planning quality even with a single iteration; performance reaches its optimal level at 2–3 iterations, but slightly degrades when the iteration count increases to 4 or more.  

Specifically, excessive iterations tend to cause two issues: first, the model may over-focus on fine-tuning local trajectory details, neglecting the rationality of the global path; second, as the number of iterations increases, the diversity of trajectory candidate sets gradually diminishes, with higher similarity among high-quality solutions, making it difficult to cover potential optimal solutions in complex scenarios. Meanwhile, cumulative errors in the computation process gradually amplify, ultimately leading to slight performance degradation.  

Considering the trade-off between planning performance and computational efficiency (excessive iterations significantly increase inference time), we select 2 iterations as the default setting.

\begin{figure}[h!]
  \centering
  \includegraphics[width=0.98\linewidth]{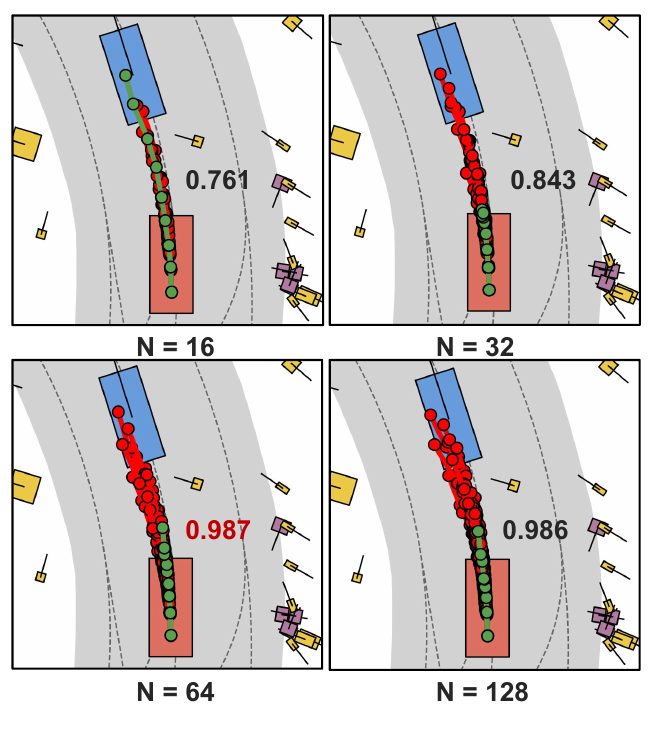}
  % \vspace{-8pt}
  \caption{Relationship between the number of anchors and trajectory scores. The scores in the figure represent the score of a single trajectory.}
  \label{fig:actor_architecture}
  % \vspace{-15pt}
\end{figure}
\subsection{Preference Weight Balancing}  
\label{sec:ablation_preference}  
Table 7 presents the experimental results of trajectory generation under different weight configurations for reward signal dimensions, verifying the model’s adaptability to diverse preference settings.  

We first attempted to parameterize the weight matrix $\{w_i\}_{i=1}^N$ and allow the model to learn it autonomously, but this approach yielded poor performance. The core issue lies in inherent flaws in the autonomous optimization: during training, the model exploits loss landscape asymmetry, prioritizing dimensions where loss reduction is computationally easy (e.g., minor tweaks to comfort or progress metrics) while neglecting critical but hard-to-optimize dimensions (e.g., safety metrics requiring complex trajectory adjustments). This short-sighted focus on immediate loss minimization leads to reward collapse—the model gradually sacrifices key characteristics to simplify optimization, degrading performance across core evaluation criteria. Fundamentally, the optimization process lacks mechanisms to ensure balanced attention to all critical dimensions, causing it to favor loss reduction shortcuts over robust performance.  

The \textbf{Conservative} setting employs a weight distribution (0.4:0.2:0.1:0.2:0.1 for NC:DAC:EP:TTC:C), prioritizing safety via higher weights for no-collision (NC) and drivable area compliance (DAC) subscores. This configuration emphasizes collision avoidance and strict adherence to drivable areas, reflecting a risk-averse strategy that places safe operation above route progress (EP).  

In contrast, the \textbf{Aggressive} configuration (0.1:0.2:0.4:0.1:0.2 for NC:DAC:EP:TTC:C) shifts focus to efficiency by significantly increasing the weight of ego progress (EP) to prioritize route advancement. This setup deliberately trades moderate reductions in safety metrics (NC, TTC) for faster route completion efficiency.  

The \textbf{Balanced} weights (0.2:0.2:0.2:0.2:0.2) distribute importance equally across all subscores (no collisions, drivable area compliance, ego progress, time-to-collision, and comfort), avoiding extreme prioritization of any single aspect and achieving stable, balanced performance across evaluation criteria.  

Notably, the \textbf{Comfort-focused} setting (0.2:0.2:0.1:0.1:0.4 for NC:DAC:EP:TTC:C) assigns the highest weight to comfort (C), explicitly prioritizing smooth driving characteristics (e.g., acceleration and jerk constraints). It retains baseline weights for safety (NC, DAC) and progress (EP) to ensure improved comfort does not come at the cost of significant sacrifices in these areas, validating the model’s adaptability to explicit comfort preferences.

\subsection{Design of Two-Stage Actor}
Table 8 demonstrates the effectiveness of the stage-wise design module in our proposed Actor. Experimental results show that using only the autoregressive module or the diffusion module as the trajectory generator, or reversing the order of the two modules, all result in inferior performance compared to our designed architecture.

\subsection{Anchor Number Configuration}
\label{sec:ablation_anchor_number}
We analyzed how anchor point count impacts trajectory quality (Table~\ref{tab:ablation_anchor_number}). The \textbf{16-anchor} setup performed poorest due to insufficient sampling density. Increasing to \textbf{32 anchors} yielded a notable PDMS jump to 92.9, with improved safety and efficiency. \textbf{64 anchors} further optimized performance (PDMS=94.9) with optimal safety and comfort, while \textbf{128 anchors} showed negligible gains, confirming 64 as the optimal balance of sampling density and computational efficiency.

%\newpage

% {
%     \small
%     %\bibliographystyle{aaai2026}
%     \bibliography{aaai2026}
% }

\end{document}